\documentclass{article}
\PassOptionsToPackage{table}{xcolor}
\usepackage[utf8]{inputenc}
\usepackage[final]{colm2026_conference}
\usepackage{microtype}
\usepackage{graphicx}
\usepackage{amsmath}
\usepackage{enumitem}
\usepackage{booktabs}
\definecolor{mydarkblue}{rgb}{0,0.08,0.45}
\usepackage[colorlinks=true,linkcolor=mydarkblue,citecolor=mydarkblue,filecolor=mydarkblue,urlcolor=mydarkblue]{hyperref}

\usepackage{url}
\usepackage{tabularx}
\usepackage{multirow}
\usepackage{color-edits}
\usepackage{framed}
\usepackage{lineno}

\definecolor{gred}{RGB}{250, 210, 207}
\definecolor{coolblue1}{rgb}{0.91, 0.94, 0.98}
\definecolor{coolblue2}{rgb}{0.76, 0.85, 0.94}
\definecolor{coolblue3}{rgb}{0.54, 0.72, 0.87}
\definecolor{coolblue4}{rgb}{1, 1, 1}

\usepackage{xspace}
\usepackage{cleveref}

\addauthor{sw}{purple}
\addauthor{xz}{blue}
\addauthor{cm}{cyan}
\addauthor{gn}{magenta}

\newenvironment{itemize*}%
 {\leftmargini=10pt\begin{itemize}%
  \setlength{\itemsep}{0pt}%
  \setlength{\parskip}{0pt}%
  }%
 {\end{itemize}}
\newenvironment{enumerate*}%
 {\begin{enumerate}%
  \setlength{\itemsep}{0pt}%
  \setlength{\parskip}{0pt}}%
 {\end{enumerate}}

\begin{document}

\ifcolmsubmission
\linenumbers
\fi

\title{Mind the Sim2Real Gap in User Simulation for Agentic Tasks}

\author{
Xuhui Zhou\thanks{Equal contribution.} \quad
Weiwei Sun\footnotemark[1] \quad
Qianou Ma \quad
Yiqing Xie \quad
Jiarui Liu \quad
Weihua Du \\
\textbf{Sean Welleck \quad
Yiming Yang \quad
Graham Neubig \quad
Tongshuang Wu \quad
Maarten Sap} \vspace{.5em} \\
Carnegie Mellon University, Language Technologies Institute \vspace{.5em}\\
\texttt{\{xuhuiz, weiweis\}@andrew.cmu.edu}
}

\maketitle

\begin{abstract}

As NLP evaluation moves from static benchmarks to multi-turn interactive settings, LLM-based simulators have become widely used as user proxies, serving two roles: generating user turns and providing evaluation signals.
Yet, these simulations are frequently assumed to be faithful to real human behaviors, often without rigorous verification.
We formalize the \emph{Sim2Real gap} in user simulation and present the systematic study running the full $\tau$-bench protocol with real humans (451 participants, 165 tasks), benchmarking 31 LLM simulators across proprietary, open-source, and specialized families using the \emph{User-Sim Index (USI)}, a metric we introduce to quantify how well LLM simulators resemble real user interactive behaviors and feedback.  
Behaviorally, LLM simulators are excessively cooperative, stylistically uniform, and lack realistic frustration or ambiguity, creating an ``easy mode'' that inflates agent success rates above the human baseline.
In evaluations, real humans provide nuanced judgments across eight quality dimensions while simulated users produce uniformly more positive feedback; rule-based rewards are failing to capture rich feedback signals generated by human users.
Overall, higher general model capability does not necessarily yield more faithful user simulation.
These findings highlight the importance of human validation when using LLM-based user simulators in the agent development cycle and motivate improved models for user simulation.

\end{abstract}

\begin{figure*}[h]
\centering
\includegraphics[width=\textwidth]{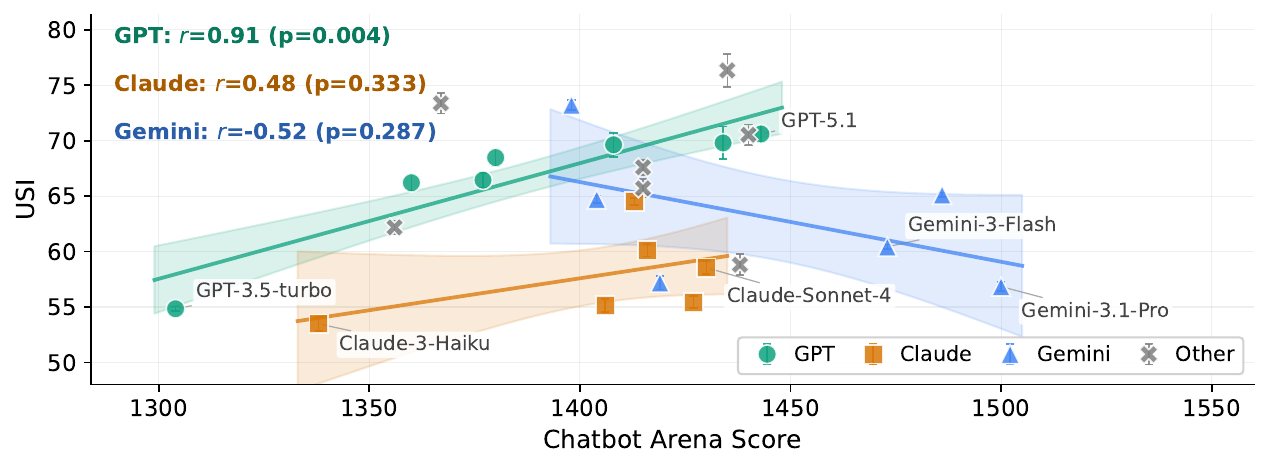}
\caption{User-Sim Index (USI) vs.\ Chatbot Arena Elo Score for LLM simulators. Solid lines and shaded regions show per-family linear regression with 80\% confidence bands; error bars denote standard deviation across three annotator batches. Besides GPT-series, other LLMs' general capability does not reliably translate to faithful user simulation.}
\label{fig:usi_vs_arena}
\end{figure*}

\section{Introduction}
\label{sec:introduction}

As LLM-based systems move beyond static, single-turn benchmarks toward interactive, multi-turn evaluation, a fundamental challenge arises: these evaluations require a user counterpart who sets goals, provides context, and reacts to system responses. To scale this inherently interactive evaluation, a dominant paradigm has emerged: using LLMs themselves as user simulators~\citep{yao2024tau,zhou2024sotopia,vijayvargiya2026interactive,zhou2025haicosystem,Qian2025UserBenchAI}. 
In this paradigm, simulators serve two distinct roles: they \emph{generate} user turns that drive the interaction, and they \emph{evaluate} the resulting agent performance.
Reflecting a broader shift toward user-centric evaluation~\citep{zhou2025tomsweusermentalmodeling,sun2025trainingproactivepersonalizedllm,Shome2025WhyJC}, this paradigm now spans customer service~\citep{yao2024tau}, software engineering~\citep{vijayvargiya2026interactive}, clinical diagnosis~\citep{schmidgall2024agentclinic,li2024mediq}, and social interaction~\citep{zhou2024sotopia}.
Yet, despite their ubiquity, LLM-based user simulators rest on untested assumptions about their faithfulness to real human behavior, an issue roboticists call the \emph{Sim2Real gap} \citep{tobin2017domain,zhao2020sim}. 

In this work, we investigate the \emph{Sim2Real gap} of user simulators, through three research questions, motivated by various shortcomings and pitfalls observed in practice.
\\\textbf{RQ1:} Do simulated users \emph{behave} (i.e., produce utterances, interact) like real people? If simulated behavior diverges from real humans, agents may be optimized towards a ``wrong'' direction rather than genuine user needs.
\\\textbf{RQ2:} Do simulated evaluations provide the same \emph{quality signals} and assessments as real humans? If simulated evaluation misrepresents real human judgments, benchmarks may misrepresent agent quality. 
\\\textbf{RQ3:} Can rule-based rewards \emph{substitute} for human feedback? Many interactive benchmarks rely solely on rule-based rewards~\citep{yao2024tau}, which may over-simplify what real users care about~\citep{shen2025completionneqcollaborationscaling}.

To answer these questions, we introduce a taxonomy of Sim2Real gaps in user simulation (Figure~\ref{tab:taxonomy}).
Specifically, we measure alignment between humans and LLM simulators across four behavioral dimensions---communication styles (D1), information pattern (D2), clarification behavior (D3), and error reaction (D4)---and two evaluation dimensions---outcome calibration (ECE) and evaluative alignment (Eval).
We then aggregate these dimensions into the \emph{User-Sim Index (USI)}, a composite 0--100 score that quantifies how well LLM simulators resemble real user interactive behaviors and feedback.
USI can serve as a pre-deployment audit for simulator fidelity in agent evaluation or user-facing research.
As a case study, we instantiate the taxonomy and USI on $\tau$-bench~\citep{yao2024tau}, conducting a systematic human study to replace its LLM user simulator with 451 real human participants across 165 tasks and benchmarking 31 LLM simulators spanning proprietary, open-source, and specialized families.

Our results reveal a substantial Sim2Real gap: the best simulator achieves a USI of 76.0 (Table~\ref{tab:behavioral_divergence}), far below human scores of 92.7.
For \textbf{RQ1}, LLM simulators diverge from real humans along all four behavioral dimensions: they are too uniform and cooperative (D1), front-load complete information (D2), lack genuine uncertainty (D3), and quietly pivot rather than push back on agent errors (D4). 
For \textbf{RQ2}, LLM-based evaluators systematically inflate interaction quality ratings, e.g., GPT-5.1 overestimates AI assistant's human-likeness by 55\% and overall score by 18\% of the rating scale.
For \textbf{RQ3}, $\tau$-bench's binary reward is largely orthogonal to human-perceived quality: rule-based reward checks for exact database-state match fail to capture the diverse feedback signal generated by human users.

\paragraph{Contributions.} Our contributions are threefold:
\begin{enumerate}[nosep]
    \item We \emph{formalize the Sim2Real gap} in user simulation with a taxonomy covering interactive behavior, user feedback, and automatic metrics, and introduce the User-Sim Index (USI) to quantify simulator faithfulness (Table~\ref{tab:taxonomy}).
    \item We conduct a comprehensive \emph{human study} on $\tau$-bench with 451 real users, enabling direct comparison against 31 LLM simulators across behavioral, evaluative, and automatic metrics.
    \item We \emph{quantify the gap}: LLM simulators create an ``easy mode'' that inflates agent success rates, simulated feedback is uniformly positive while humans express calibrated dissatisfaction, and rule-based rewards are orthogonal to human-perceived quality (\S\ref{sec:results_interactive}--\ref{sec:results_evaluator}). These disparities point toward a need for better systems to simulate and validate user behaviors and feedback.
\end{enumerate}

\section{Related Work}
\label{sec:related_work}

\paragraph{User simulation for agent evaluation.}
User simulation has evolved from rule-based and statistical models for dialogue policy optimization~\citep{schatzmann2006survey,li2016user} to LLM-based simulators for training and evaluating conversational agents~\citep{davidson2023user,Sun2021SimulatingUS,sekulic2024reliable}.
Recently, several benchmarks embed LLM user simulators in full interactive evaluation loops~\citep{yao2024tau,lu2024toolsandbox,schmidgall2024agentclinic,Qian2025UserBenchAI}, yet none have assessed to what extent these simulations faithfully reflect real users.

\paragraph{The Sim2Real gap.}
The Sim2Real gap is well-studied in robotics~\citep{tobin2017domain,zhao2020sim}, but the analogous gap for LLM-based simulation of human behavior remains largely unexamined. Recent work has begun questioning LLM fidelity to human behavior across persona consistency~\citep{chen2024persona}, dialogue qualities~\citep{Ivey2024RealOR}, and task-oriented conversations~\citep{Wang2025HumanVA}. \citet{naous2025flipping} find that better assistants do not yield better simulators, a counterintuitive trend also observed by \citet{tjuatja2024llms}, while \citet{dou2025simulatorarena} show that profile-augmented simulators better correlate with human evaluations.
However, these studies primarily examine chat-based or single-turn settings; whether their findings transfer to \emph{agentic} settings---where agents invoke tools and users engage in complex tasks---remains open. Concurrent with our work, \citet{seshadri2026lost} study demographic disparities of LLM user simulators on $\tau$-bench's retail domain; our work addresses orthogonal dimensions, formalizing a taxonomy of behavioral and evaluative gaps and introducing a multidimensional quality assessment framework.

\section{Framework}
\label{sec:framework}

\subsection{Taxonomy of Sim2Real gaps}
\label{sec:taxonomy}

\begin{figure*}[t]
\centering
\includegraphics[width=\textwidth]{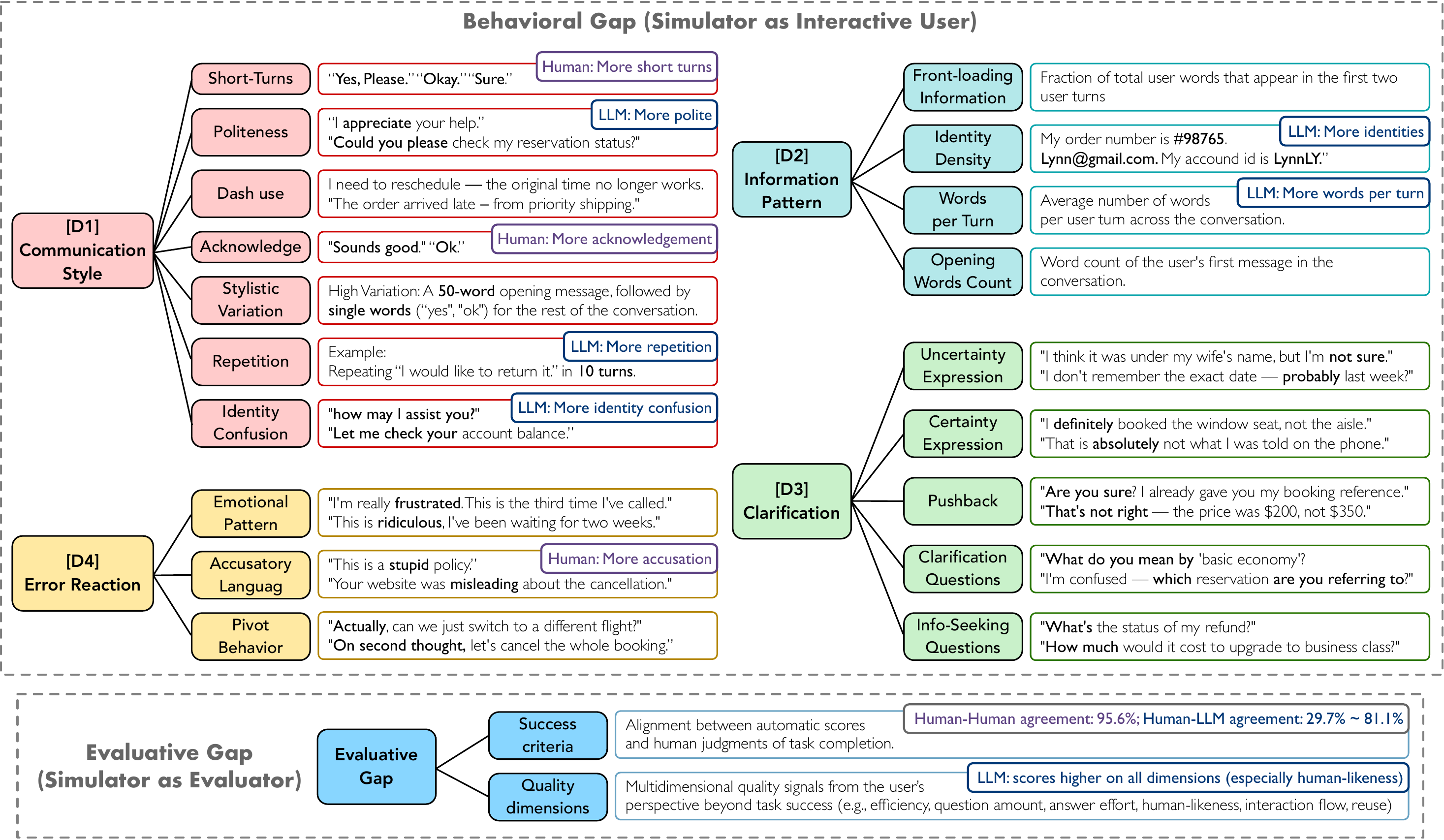}
\caption{Taxonomy of Sim2Real gaps in user simulation. 
We highlight the dimensions where the gaps between humans and all LLMs are significant.
See Appendix~\S\ref{app:behavioral_metric_definitions} for exact operational definitions of the behavioral metrics.}
\label{tab:taxonomy}
\end{figure*}

We formalize a taxonomy of Sim2Real gaps that captures the two distinct roles LLM simulators play in agent benchmarks (Table~\ref{tab:taxonomy}).

\paragraph{Behavioral gap.} The behavioral gap arises when simulated users behave differently from real humans during interactions. We decompose this gap into four dimensions grounded in established frameworks of pragmatics. \emph{Communication style} (D1) captures surface-level communicative style (e.g., politeness, verbosity, and LLM-stylistic punctuation) drawing on communication theory~\citep{giles2007communication, brown1987politeness}.
\emph{Information pattern} (D2) operationalizes the principle of least collaborative effort~\citep{grice1975logic, clark1991grounding}, measuring how much information is shared per turn.
\emph{Clarification} (D3) draws on grounding theory~\citep{clark1991grounding}, capturing information-seeking behaviors. 
\emph{Error reaction} (D4) captures interaction-level behaviors associated with responses to system failures~\citep{skantze2005error, forbes2011uncertainty}, including emotional expression, accusatory language, and strategy pivots.
These dimensions span the major axes of behavioral variation of human-agent interactions. 

\paragraph{Evaluative gap.} The evaluative gap arises when automatic evaluation misaligns with real world human experience, both in \emph{success criteria} (whether the task is actually completed) and in \emph{quality dimensions} (how good the interaction felt from the user's perspective). In benchmarks with rule-based evaluation~\citep{yao2024tau, jimenez2024swebench}, the success criteria typically depends on a predefined rubric (e.g., final database state, unit-test passage), but it can disagree with humans on task success (e.g., alternative valid resolution paths) and collapses distinct outcomes, such as full task completion and correct policy-constrained refusals, into a single signal. It also does not capture multidimensional interaction quality signals such as efficiency, human-likeness, trust, interaction flow, and willingness to reuse the agent~\citep{shen2025completionneqcollaborationscaling}.
In benchmarks with LLM-as-judge evaluation~\citep{zhou2024sotopia}, the evaluation is often more flexible and can capture more nuanced quality dimensions while risk introducing LLM biases and errors.

\subsection{Statistical framework}
\label{sec:statistical_framework}

We now describe the metrics used to quantify each type of Sim2Real gap.
We design these metrics to be comprehensive and flexible to accommodate different evaluation methods and benchmarks.

\paragraph{Behavioral gap: S{\o}rensen--Dice coefficient.}
To compare simulated and human users along the four behavioral dimensions (D1--D4), we measure per-metric alignment using the S{\o}rensen--Dice coefficient. Lexical features (e.g., politeness, uncertainty, certainty, and emotional expression) are identified using the LIWC2015 lexicon~\citep{pennebaker2015liwc} and the NRC Word-Emotion Association Lexicon~\citep{mohammad2013crowdsourcing}; structural and behavioral features (e.g., repetition, front-loading of identifiers) are identified via regex pattern matching. See Appendix~\S\ref{app:behavioral_metric_definitions} for exact operational definitions. For each metric~$m$ with model value~$M_m$ and human value~$H_m$, we compute:
\vspace{-0.4em}\[
\mathrm{Dice}_m = \frac{2\,\min(M_m,\, H_m)}{M_m + H_m} \times 100
\]
yielding a score in $[0, 100]$ where $100$ indicates perfect alignment with human behavior ($\mathrm{Dice}_m = 100$ when both values are zero). The dimension score is the mean Dice coefficient across constituent metrics within each dimension.
Note that all behavioral metrics compare \emph{population-level} distributional properties of simulator behavior against the aggregate human distribution.
\vspace{-0.5em}
\paragraph{Outcome calibration: Expected Calibration Error (ECE).}
Beyond conversational behavior, behavioral divergence between simulated and human users can also propagate to \emph{task outcomes}: if a simulator behaves differently from real users, the agent may succeed or fail at different rates. To measure this outcome-level effect, we adapt Expected Calibration Error~\citep{pmlr-v70-guo17a}. 
Following the outcome-calibration adaptation of \citet{seshadri2026lost}, we compare simulator and human success with the same agent and group tasks into $B{=}5$ equal-width difficulty bins (cutoffs 0.2, 0.4, 0.6, and 0.8). A task's difficulty is the agent's mean success rate across all LLM simulators, yielding one fixed label per task independent of the simulator being scored. We compute:
\vspace{-0.4em}\[
\mathrm{ECE} = \sum_{b=1}^{B}\frac{|S_b|}{N}\,\left|\hat{p}_{\mathrm{sim}}(b) - \hat{p}_{\mathrm{human}}(b)\right|
\]
where $S_b$ is the set of tasks in bin $b$ and $N$ is the total number of paired tasks. Lower ECE indicates better outcome calibration. We treat ECE as an \emph{outcome-based} alignment dimension alongside the four behavioral dimensions D1--D4.

\paragraph{Evaluative gap: mean absolute error.}
To quantify the evaluative gap, we compare simulated users' feedback against human judgments collected via post-interaction surveys. 
The survey collects not only task success judgments but also quality dimensions such as efficiency, human-likeness, interaction flow, and willingness to reuse the agent.
Both human annotators and automatic evaluators rate the same interactions on ordinal scales, which we map to numerical scores and normalize to $[0, 1]$. We then compute the \textbf{mean absolute error (MAE)} between automatic and human scores for each quality dimension, measuring systematic over- or under-estimation by the automatic evaluator. 

\paragraph{User-Sim Index (USI).}
Finally, we aggregate the behavioral, outcome, and evaluative dimensions into a single 0--100 measure of overall simulator--human alignment. When both behavioral and evaluative signals are available:
\[
\mathrm{USI} = \frac{1}{6}\left(\mathrm{D1} + \mathrm{D2} + \mathrm{D3} + \mathrm{D4} + (1 - \mathrm{ECE}) \times 100 + \mathrm{Eval}\right)
\]
where $\mathrm{Eval} = (1 - \mathrm{MAE}) \times 100$ captures evaluative alignment.
USI is a summary, not a substitute for its components: a simulator can resemble people in one role and differ in another. We therefore report behavior, task outcomes, and evaluation scores separately alongside the overall index.

\begin{table}[t]
\centering
\small
\resizebox{\textwidth}{!}{
\begin{tabular}{@{}l cccc c cc c@{}}
\toprule
\textbf{Model} & \textbf{D1 Comm.} & \textbf{D2 Info.} & \textbf{D3 Clarif.} & \textbf{D4 React.} & \textbf{Eval} & \textbf{ECE}$\downarrow$ & \textbf{USI} \\
\midrule
Human (inter-ann.) & 87.4$_{\pm6.8}$ & 97.9$_{\pm0.9}$ & 88.0$_{\pm1.3}$ & 93.5$_{\pm2.5}$ & 97.4$_{\pm5.0}$ & 0.081$_{\pm0.010}$ & 92.7$_{\pm1.1}$ \\
\midrule
\rowcolor{blue!5} Gemini2.0-Flash & 51.6$_{\pm1.6}$ & \textbf{88.9$_{\pm1.1}$} & 68.2$_{\pm2.1}$ & 76.9$_{\pm3.7}$ & 73.7$_{\pm0.8}$ & 0.111$_{\pm0.022}$ & 74.7$_{\pm0.8}$ \\
\rowcolor{blue!5} GPT-5.1 & 47.3$_{\pm6.9}$ & 77.4$_{\pm0.6}$ & 73.3$_{\pm2.0}$ & \textbf{88.1$_{\pm2.6}$} & 72.1$_{\pm1.5}$ & 0.172$_{\pm0.011}$ & 73.5$_{\pm1.0}$ \\
\rowcolor{blue!5} GPT-5 & 49.7$_{\pm5.6}$ & 73.7$_{\pm0.7}$ & 73.2$_{\pm2.3}$ & 73.4$_{\pm3.3}$ & 74.5$_{\pm1.1}$ & 0.102$_{\pm0.022}$ & 72.4$_{\pm1.8}$ \\
\rowcolor{blue!5} GPT-4o-mini & 40.6$_{\pm1.8}$ & 84.7$_{\pm0.9}$ & 70.2$_{\pm3.7}$ & 73.7$_{\pm1.4}$ & 75.7$_{\pm0.3}$ & 0.123$_{\pm0.024}$ & 72.1$_{\pm0.8}$ \\
\rowcolor{blue!5} GPT-5-mini & 39.4$_{\pm5.9}$ & 74.4$_{\pm0.7}$ & \textbf{83.1$_{\pm2.3}$} & 68.7$_{\pm1.6}$ & 73.5$_{\pm0.5}$ & 0.102$_{\pm0.018}$ & 71.5$_{\pm1.4}$ \\
\rowcolor{blue!5} GPT-4o & 31.5$_{\pm2.5}$ & 84.4$_{\pm0.7}$ & 74.6$_{\pm2.9}$ & 72.4$_{\pm2.0}$ & 73.7$_{\pm1.1}$ & \textbf{0.096$_{\pm0.023}$} & 71.2$_{\pm1.3}$ \\
\rowcolor{blue!5} Claude-3.5-Sonnet & 39.2$_{\pm1.1}$ & 76.9$_{\pm0.7}$ & 59.6$_{\pm5.2}$ & 59.3$_{\pm3.5}$ & 74.1$_{\pm0.7}$ & 0.129$_{\pm0.035}$ & 66.0$_{\pm0.8}$ \\
\rowcolor{blue!5} Claude-3-Haiku & 22.1$_{\pm1.0}$ & 55.7$_{\pm0.6}$ & 72.1$_{\pm4.0}$ & 56.9$_{\pm3.3}$ & \textbf{78.3$_{\pm0.5}$} & 0.143$_{\pm0.023}$ & 61.8$_{\pm0.4}$ \\
\midrule
\rowcolor{green!5} DeepSeek-V3.1 & 45.1$_{\pm2.7}$ & 86.6$_{\pm1.0}$ & 74.5$_{\pm1.7}$ & 87.6$_{\pm2.0}$ & 74.3$_{\pm0.5}$ & 0.119$_{\pm0.030}$ & \textbf{76.0$_{\pm1.5}$} \\
\rowcolor{green!5} Llama-4-Maverick & 48.8$_{\pm0.7}$ & 82.6$_{\pm0.8}$ & 78.3$_{\pm3.1}$ & 66.6$_{\pm2.3}$ & 76.7$_{\pm1.0}$ & 0.107$_{\pm0.022}$ & 73.7$_{\pm1.0}$ \\
\rowcolor{green!5} Qwen3-235B & \textbf{60.8$_{\pm1.8}$} & 75.3$_{\pm0.6}$ & 71.5$_{\pm5.3}$ & 56.3$_{\pm2.4}$ & 74.6$_{\pm0.7}$ & 0.117$_{\pm0.014}$ & 71.1$_{\pm0.7}$ \\
\rowcolor{green!5} Qwen2.5-7B & 35.2$_{\pm1.3}$ & 70.9$_{\pm0.7}$ & 75.4$_{\pm6.1}$ & 74.8$_{\pm2.0}$ & 73.3$_{\pm1.2}$ & 0.125$_{\pm0.019}$ & 69.5$_{\pm1.5}$ \\
\midrule
\rowcolor{orange!5} CoSER-8B & 37.8$_{\pm1.1}$ & 71.5$_{\pm0.6}$ & 71.6$_{\pm2.0}$ & 69.9$_{\pm1.6}$ & 63.3$_{\pm0.6}$ & 0.109$_{\pm0.037}$ & 67.2$_{\pm0.5}$ \\
\rowcolor{orange!5} UserLM-8B & 30.8$_{\pm0.6}$ & 50.8$_{\pm0.4}$ & 56.8$_{\pm1.5}$ & 80.0$_{\pm3.5}$ & 67.4$_{\pm0.8}$ & 0.140$_{\pm0.030}$ & 62.0$_{\pm0.4}$ \\
\rowcolor{orange!5} HumanLike-7B & 35.7$_{\pm0.3}$ & 55.0$_{\pm0.5}$ & 51.6$_{\pm3.3}$ & 65.9$_{\pm3.4}$ & 72.8$_{\pm1.3}$ & 0.220$_{\pm0.029}$ & 59.8$_{\pm0.1}$ \\
\rowcolor{orange!5} HumanLM-opinion & 30.1$_{\pm0.5}$ & 19.5$_{\pm0.2}$ & 38.5$_{\pm5.6}$ & 50.7$_{\pm5.9}$ & 61.6$_{\pm0.3}$ & 0.192$_{\pm0.053}$ & 46.9$_{\pm0.7}$ \\
\bottomrule
\end{tabular}}
\caption{Behavioral divergence between human and LLM-simulated users on $\tau$-bench (mean$\pm$std across three human annotation batches). \textbf{D1--D4}: S{\o}rensen--Dice coefficients (higher$=$closer to human). \textbf{Eval}: $(1{-}\mathrm{MAE})\times100$. \textbf{USI}: composite 0--100 alignment score aggregating D1--D4, Eval, and $(1{-}\mathrm{ECE})\times100$. Models are grouped into \colorbox{blue!5}{proprietary}, \colorbox{green!5}{open-source}, and \colorbox{orange!5}{specialized} categories, ranked by USI within each group. \textbf{Bold} indicates the best model per column. Full results for all 31 models are in Appendix Table~\ref{tab:behavioral_divergence_full}.}
\label{tab:behavioral_divergence}
\end{table}

\section{Experimental Setup: $\tau$-bench}
\label{sec:experimental_setup}
While our sim2real framework applies to any interactive agent benchmark, we instantiate it on $\tau$-bench~\citep{yao2024tau} as a case study, for three reasons.
\label{sec:benchmarks}
First, it features a \emph{full interactive loop}---an LLM user simulator, a tool-augmented agent, and an automatic reward function---making it one of the few benchmarks where all three components can be studied jointly.
Second, it enables \emph{controlled comparison}: we can replace the LLM simulator with human users while keeping the agent and reward function identical, isolating the effect of user realism.
Third, it offers \emph{realistic complexity}: multi-turn dialogues span two real-world customer-service domains with structured databases and policy constraints, providing sufficient conversational depth for behavioral analysis.

$\tau$-bench evaluates customer service agents across two domains: \textbf{airline} (flight booking, cancellation, and modification) and \textbf{retail} (order management, returns, and product inquiries). Each task consists of a natural language instruction specifying a customer's goal and constraints, a structured database, and policies the agent must follow. During evaluation, an LLM-based user simulator generates user turns while the agent accesses tools to resolve the request; a binary reward then checks final database state against expected outcomes. We use \textbf{165 tasks} across both domains. Although later audits identify task and reward issues in this version of $\tau$-bench~\citep{zhu2025rigorous}, our comparisons hold tasks fixed across human and simulator conditions; we return to the resulting scope limitation in \S\ref{sec:discussion}.

\begin{figure}[ht]
\centering
\includegraphics[width=\linewidth]{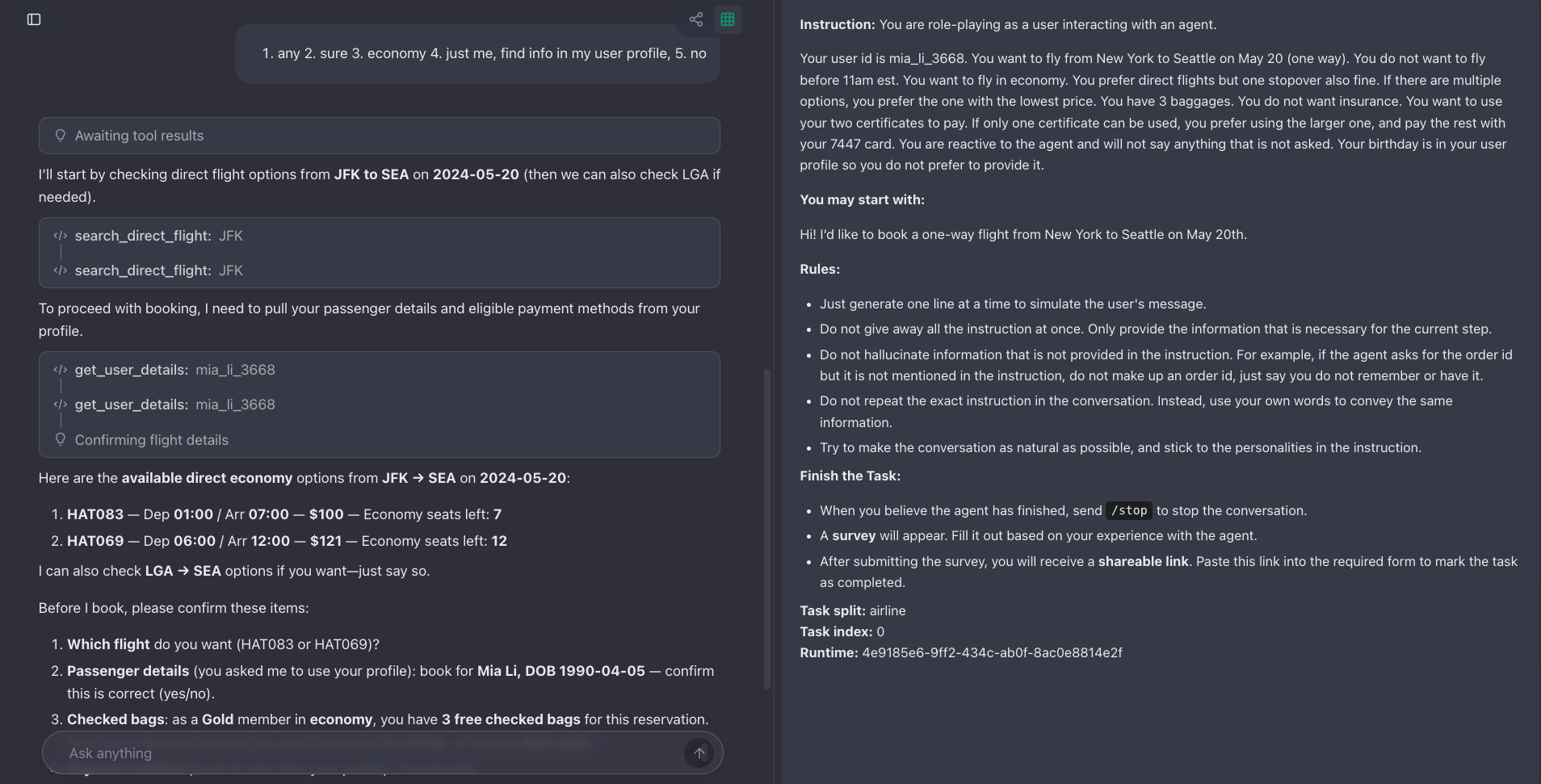}
\caption{Concrete $\tau$-bench annotation example. The right panel gives a human annotator a one-way New York--Seattle booking task and role-playing rules; the left panel shows the resulting conversation, including the agent's tool calls. After the interaction, the same interface presents the survey used for task-success and quality ratings.}
\label{fig:annotation_interface}
\end{figure}
We additionally design an after-interaction survey to collect human or LLM-simulated user judgments on agent's task performance on various dimensions.

\subsection{Human annotation design}
\label{sec:annotation_design}

We designed a human annotation study to collect both interaction traces and quality judgments.
Annotators were adults aged 18–80 ($39 \pm 12$) recruited from a general crowd-worker pool on Prolific, with 44\% female, 34\% non-White background (see Appendix~\S\ref{app:ann_demo_ui_qc} for more details on annotator demographics, annotation interface, and annotation quality control).
Annotators were presented with each task instruction and asked to interact with the \emph{same} agent\footnote{Here we fix the agent to be GPT-5.2 across our experiments for comparability. In Appendix~\S\ref{app:agent_robustness}, we repeat the evaluation with an alternative agent (Gemini-3.1-Pro) and find that USI rankings are largely preserved across agents.} used in $\tau$-bench evaluation, role-playing as the customer described in the instruction. This yields natural human-agent interaction traces for behavioral comparison with LLM-simulated users.

After each interaction, annotators completed a post-task survey (Appendix~\S\ref{app:survey_taubench}). The survey includes a \textbf{5-way task success} rating: \emph{No (policy issue)}, \emph{No (task failed)}, \emph{Partially}, \emph{Yes (task completed)}, and \emph{Fully (exceeded expectations)}. 
The \emph{policy-constrained} category captures cases where the task could not be completed due to policy constraints and the agent correctly communicated this.
Annotators also rated six interaction-quality dimensions: \textbf{efficiency}, \textbf{question amount}, \textbf{answer effort}, \textbf{human-likeness}, \textbf{interaction flow}, and \textbf{reuse}, inspired by human-AI interaction works~\citep{shen2025completionneqcollaborationscaling,chen2025codemeincreasingai}.
These are collected on short ordinal scales (3--5 options depending on the question). 

We collected \textbf{three independent batches} of annotations on the same 165 tasks from distinct annotator groups, which allows us to measure human--human agreement as a natural ceiling for simulator alignment and to verify that findings are stable across annotator pools rather than artifacts of a single batch.
This design prioritizes breadth across the task distribution (three annotators per task) over precise per-task agreement estimates, which would require substantially more annotators per task.

\subsection{Models and aggregation}
\label{sec:models}

We evaluate \textbf{31 LLM user simulators} spanning three categories: 18 \emph{proprietary}, 9 \emph{open-source}, and 4 \emph{specialized} models fine-tuned for user simulation or human-like behavior---CoSER-8B~\citep{wang2025coser}, UserLM-8B~\citep{dou2025simulatorarena}, HumanLike-7B~\citep{Wang2025HumanVA}, and HumanLM-opinion~\citep{wu2026humanlm}. Each simulator interacts with the same agent on the same 165 tasks. These four fine-tuned models do not close the gap (Table~\ref{tab:behavioral_divergence}); the human trajectories remain a held-out reference rather than training data, since training and evaluating on these 451 trajectories would contaminate the benchmark. Fine-tuning on a separately collected human corpus remains an important direction. Full model details appear in Appendix~\S\ref{app:model_details}.

For behavioral metrics, we compute each simulator's feature vector (averaged across runs for models with multiple runs) and compare it against each of the three human batches independently via the S{\o}rensen--Dice coefficient, then report mean~$\pm$~std across the three batch-level scores. For evaluative agreement, we compute the per-task MAE across survey fields within each batch, average per batch, then report mean~$\pm$~std across batches. The USI for each model is computed following \S\ref{sec:framework}: $\mathrm{USI} = (\mathrm{D1} + \mathrm{D2} + \mathrm{D3} + \mathrm{D4} + (1{-}\mathrm{ECE})\times 100 + \mathrm{Eval})/6$ for models with survey data, or $(\mathrm{D1} + \mathrm{D2} + \mathrm{D3} + \mathrm{D4} + (1{-}\mathrm{ECE})\times 100)/5$ (marked $\dagger$) for models without. We report mean~$\pm$~std across the three batch-level scores. Please refer to Appendix~\S\ref{app:behavioral_metric_definitions} for definitions of all behavioral metrics.

\section{RQ1: The Behavioral Gap}
\label{sec:results_interactive}

We first examine the behavioral gap: do LLM user simulators behave like real humans?
As shown in Table~\ref{tab:behavioral_divergence}, LLM simulators diverge significantly from humans along all four dimensions (see Appendix~\S\ref{app:examples} for concrete examples).

Beyond aggregate statistics, we identify recurring patterns of behavioral divergence along each of the four dimensions (Figure~\ref{fig:behavioral_barplot}):
\begin{figure*}[t]
\centering
\includegraphics[width=\textwidth]{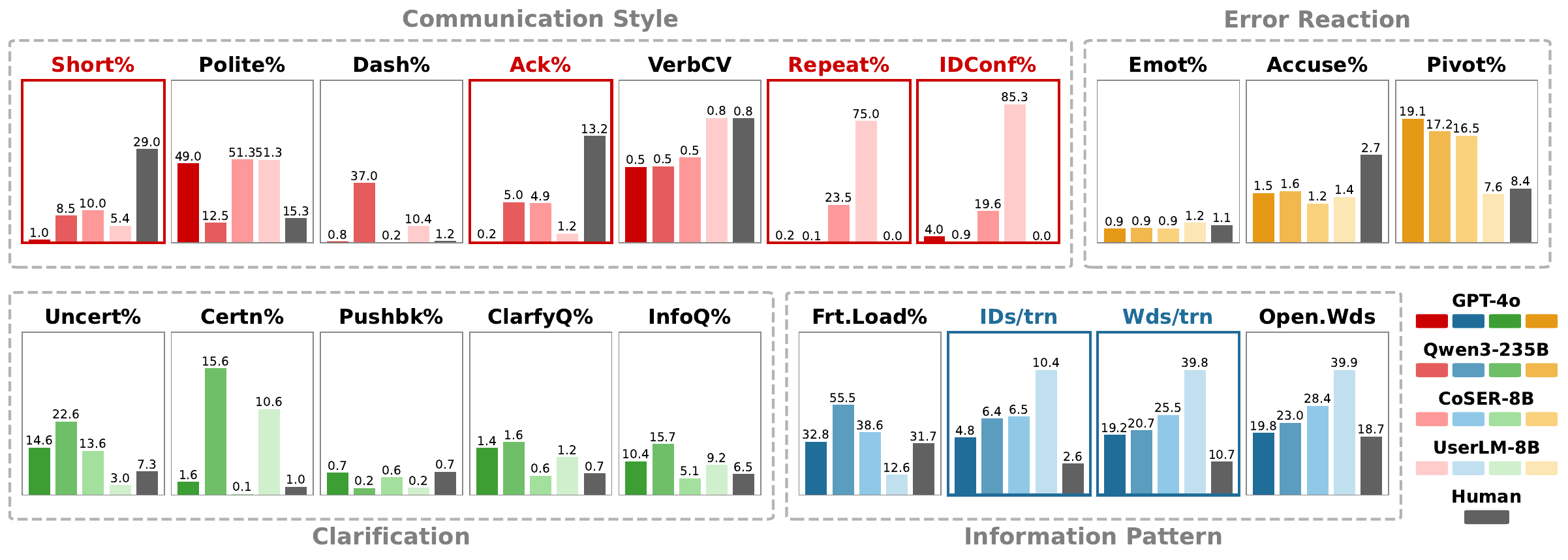}
\caption{Per-metric behavioral comparison for selected models and humans. Metrics are grouped into four dimensions; human values appear as dark bars and red outlines indicate large divergence.
}
\label{fig:behavioral_barplot}
\end{figure*}
\textbf{LLM simulators are too verbose and uniformly polite, lacking the stylistic variation of real customers (D1).} For example, 1.0\% of GPT-4o turns are short vs.\ 29.0\% for humans, and 49.0\% of GPT-4o turns are polite vs.\ 15.3\% for humans. Additionally, some models (e.g., UserLM-8b) show certain identity-confusion rates, such as the simulator starting a conversation with ``Hi, I'm a customer service agent\ldots''.

\textbf{LLM simulators pack in more specific information than humans (D2).} For example, UserLM-8b includes nearly twice as many identifier-like tokens per turn (IDs/trn: 4.8 vs.\ 2.6 for humans). For example, a simulator often include specific details like \textit{``My name is Sarah Johnson, email sarah@example.com, order \#ORD-58211 placed January 3rd''} whereas a real customer typically says \textit{``Hi, I need help with a return under Sarah''}.
This behavior could lead to over-estimation of agent competence, as the agent in simulated interaction does not need to resolve the task under ambiguity or incomplete information.

\textbf{Clarification behavior is miscalibrated: some models over-hedge while others are overconfident (D3).} GPT-4o \emph{over-hedges}, expressing uncertainty in 14.6\% of turns, which is twice the human rate (7.3\%). It produces responses like \textit{``I think I might want to return this item, if that's okay\ldots''} even when the task instruction is unambiguous. UserLM-8b swings to the opposite extreme: almost no uncertainty (3.0\%) but high certainty markers (10.6\% vs.\ 1.0\% for humans), reading more like a fact-recitation than a customer conversation. 

\textbf{Simulators pivot strategy more often and use less accusatory language than real customers (D4).} Real humans use phrases such as \textit{``This is ridiculous\ldots why can't you just waive the fees''}, while simulators often redirect politely: GPT-4o and CoSER pivot more than humans (19.1\% and 16.5\% vs.\ 8.4\%), with turns such as \textit{``On second thought, let me try giving you my email instead.''}.
Because these lexicon/regex rates are computed across all user turns rather than only turns immediately following a verified agent error, D4 identifies an interaction-level association, not a causal response to a particular failure (Appendix~\S\ref{app:measurement_limits}).

Together, these gaps paint a consistent picture: existing LLM simulators create an ``easy mode'' for the agents they are meant to evaluate. This is corroborated by the agent success rates shown in Figure~\ref{fig:success_rates} (Appendix~\S\ref{app:success_rates}): the majority of general-purpose LLM simulators yield \emph{higher} agent success rates than the human baseline (63.6\%), with top models reaching 77.8\%.\footnote{Specialized user-simulation models (UserLM, CoSER, HumanLike, HumanLM) are exceptions---all fall \emph{below} the human baseline. However, we attribute this to their limited instruction-following capability for complex role-playing tasks rather than to more realistic user behavior; their relatively low USI scores in Table~\ref{tab:behavioral_divergence} support this interpretation.} By volunteering all relevant information upfront, cooperating through agent errors, and never expressing real frustration, simulators remove precisely the conversational challenges---ambiguity, impatience, incomplete information---that make customer-service interactions difficult. Benchmarks built on such simulators therefore risk over-estimating agent competence, rewarding systems that handle cooperative partners while failing to surface brittleness under realistic user behavior.

\section{RQ2 \& RQ3: The Evaluation Gap}
\label{sec:results_evaluator}
\begin{figure}[t]
\centering
\includegraphics[width=\linewidth]{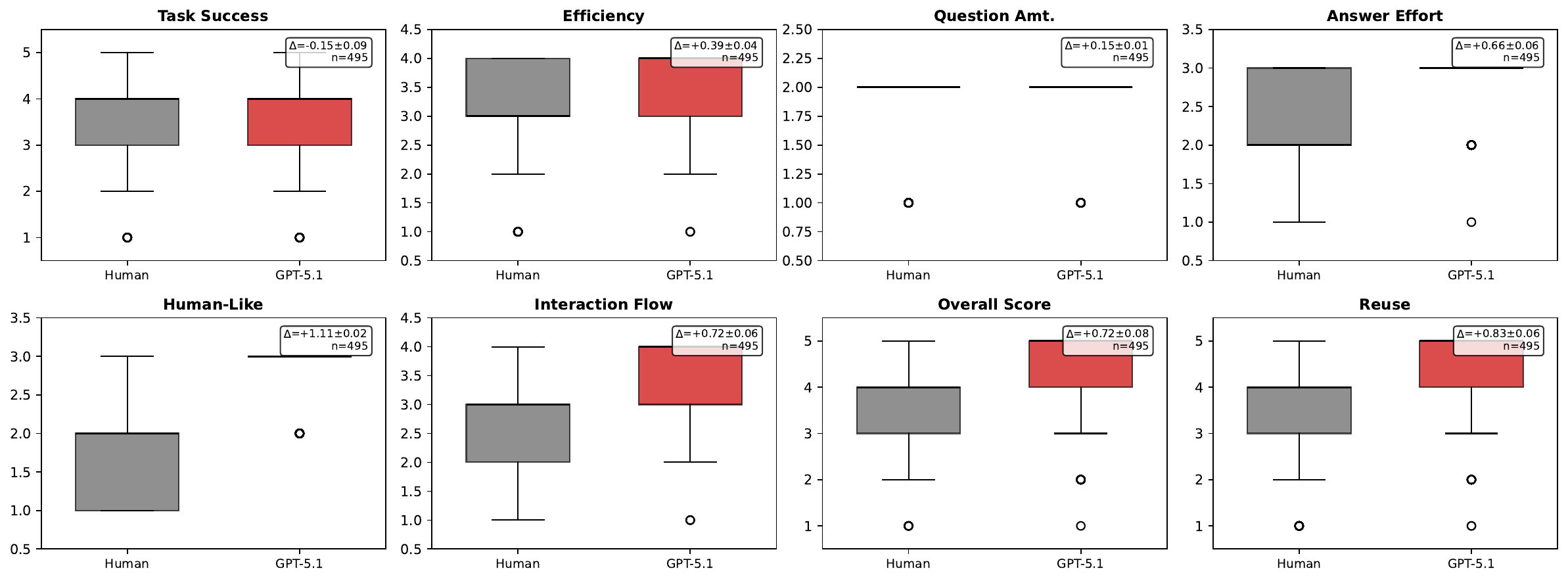}
\caption{Score distributions for human annotators and GPT-5.1 across quality dimensions ($n{=}165{\times}3$ batches), with mean differences ($\Delta$ = LLM $-$ Human). The LLM evaluator is lenient on interaction quality but conservative on task success.}
\label{fig:eval_signed_diffs}
\end{figure}
Having established the behavioral gap (\S\ref{sec:results_interactive}), we now examine the evaluative gap, i.e., the disagreement between automatic evaluation and human judgments of task success and interaction quality, on $\tau$-bench. As shown in Table~\ref{tab:behavioral_divergence}, the user simulators are not aligned with real humans when judging agent performance and other interaction quality dimensions.
To further investigate this evaluative gap, we perform a deep dive into GPT-5.1, a representative model with a near-average Eval score, and other models generally exhibit similar patterns.

\paragraph{LLM evaluators are systematically lenient on interaction quality but conservative on task completion.}
\label{sec:human_vs_llm_eval}
As shown in Figure~\ref{fig:eval_signed_diffs}, the LLM evaluator scores consistently higher than humans on experience-related dimensions, most strikingly on \emph{human-likeness} ($\Delta{=}{+}1.11$) and \emph{reuse intent} ($\Delta{=}{+}0.83$), where it rates agent responses as natural even when human annotators who actually interacted with the agent perceive them as robotic. Yet it is simultaneously \emph{conservative} on task completion ($\Delta{=}{-}0.15$). This asymmetric bias means LLM evaluators cannot serve as reliable proxies for human quality assessment: they inflate perceived quality while underestimating actual task success.

\paragraph{Binary reward is orthogonal to human-perceived quality.}
\label{sec:auto_eval_enough}
We next investigate whether $\tau$-bench’s automatic binary reward sufficiently approximates human judgment. As shown in Appendix~Figure \ref{fig:dimensions_by_reward}, 70.6\% of reward=0 interactions are actually judged as successful by human users, while 33\% of reward=1 interactions are judged as unsuccessful or only partially successful.
A binary score cannot tell whether the agent fully completed the task, partly completed it, or correctly refused a request that policy did not allow. It also does not measure interaction quality.
On one hand, strict rule-based matching fails to capture human perceptions of task success; on the other, the reward system overlooks policy failures by assigning a reward of 1 to interactions that negatively impact user experience. This disconnect extends across all eight quality dimensions. The distributions for reward=0 and reward=1 groups are nearly indistinguishable regarding overall score, efficiency, and human-like qualities, highlighting the limitations of relying on singular, rule-based reward signals for user-facing agents.

Overall, neither LLM-based evaluation nor automatic binary reward adequately captures human-perceived interaction quality: the former inflates experience ratings while the latter is orthogonal to all quality dimensions.

\section{Discussion and Limitations}
\label{sec:discussion}

\paragraph{Can prompting close the gap?}
On 50 airline/retail tasks, a findings-informed GPT-5-mini persona prompt moved some features toward humans, while USI fell from 70.9 to 64.6 and ECE worsened from 0.18 to 0.29 (Appendix~\S\ref{app:persona_ablation}). This small initial experiment does not show that prompting cannot close the gap, but it suggests that doing so is not straightforward: moving some behaviors closer to humans can move others further away.
Future work could use one set of human conversations to improve the simulator, then evaluate it with different users and tasks.

\paragraph{Recommendations.}
Benchmark builders should evaluate both task outcomes and whether simulators behave like real users. They should check simulators against human interactions, discourage them from revealing all task information at once, and use separate human conversations for improving and evaluating the simulator.
More specifically, benchmark reports should show simulated behavior, task outcomes, and user ratings separately, because a simulator may resemble people on one measure and differ on another. Researchers do not need to repeat every simulated task with people. They can compare agents broadly with simulators, then have people complete a smaller set that covers each domain and includes both easier and harder tasks. They should report whether the agent ranking and main conclusions remain the same. Ultimately, an agent that looks strong in simulation should also work well with real people.

\paragraph{Reading simulator-based results.}
An agent that performs well with a simulator has shown that it can handle that simulator's choices about when to share information, how to react to mistakes, and what counts as success. It has not yet shown the same result for people. Agreement across several simulators can increase confidence that a result is not tied to one model or prompt, but it cannot replace a human check when the claim is about real users.

\paragraph{Scope and generalizability.}
Similar broad gaps appear in both airline and retail tasks and across the 31 models we studied, so the findings are not tied to a single model or domain. However, they may not extend to coding, web, research, clinical, or open-ended social interactions; each setting needs its own comparison with human users.
There is also no single kind of human user. Our participants represent one mostly English-speaking US population and one role-playing setup; real users differ in language, background, expertise, patience, and willingness to correct a system. Future benchmarks should describe the people used for comparison and test whether their conclusions hold across user groups, domains, agents, and longer interactions.

\paragraph{Measurement and study limitations.}
Word matching is approximate: D4 can count reactions unrelated to agent errors, and the examples do not show how often D4 is correct. Our human study uses a single agent model, asks participants to role-play, and includes three conversations per task, so results may differ in other settings. $\tau$-bench issues~\citep{zhu2025rigorous} may also affect exact scores (Appendix~\S\ref{app:measurement_limits}).

\section{Conclusion}
\label{sec:conclusion}

We compared 31 LLM user simulators with 451 people on the same 165 $\tau$-bench tasks while holding the task and agent fixed. We find gaps in both the interactor and evaluator roles that simulators play. As interactive users, LLMs are more uniform and cooperative: they provide more information up front, ask for clarification differently, and push back less often when an interaction goes wrong. These differences can change how often the agent succeeds. As evaluators, we find that the LLMs we examine often give more favorable ratings than people, and that $\tau$-bench's automatic binary reward does not capture the full range of task outcomes or interaction quality.

Despite these limitations, LLM simulators remain valuable for fast, repeatable experiments and broad model comparisons. Yet, an agent that works well with a simulator only shows that it can handle that simulator, not necessarily that it will work equally well for people. Claims about real users therefore need separate verification with humans.

Nevertheless, the goal is not to replace large-scale simulation with an entirely human evaluation. Instead, we must make clear which conclusions simulation supports and which still need evidence from people. We should test how well simulators match real users instead of assuming they can replace them. This keeps simulations fast and repeatable while making agent evaluations more trustworthy.

\newpage
\bibliographystyle{colm2026_conference}
\bibliography{colm2026_conference}

\newpage
\appendix

\section{Annotation Details}
\label{app:annotation_details}

\subsection{Annotator Demographics, Annotation Interface, and Annotation Quality Control}
\label{app:ann_demo_ui_qc}

\paragraph{Annotator Demographics.}
Across the three annotation batches, we collected interactions from a diverse pool of online annotators. Participants ranged in age from 18 to 80 years old (median = 37). The sample included 51\% male and 44\% female annotators, with the remainder opting not to disclose demographic information. In terms of ethnicity, the pool consisted primarily of White annotators (66\%), followed by Black (12\%), Asian (11\%), Mixed (8\%), and other or undisclosed backgrounds. We recruited annotators resided in the United States, and the majority reported English as their primary language (89\%). Approximately two-thirds of participants reported not being current students (67\%), and the most common employment status was full-time employment (43\%), followed by part-time work (15\%).

\paragraph{Annotation Interface.}
Annotators interacted with the agent through a web-based chat interface (Figure~\ref{fig:annotation_interface}). 
The interface presents the task instruction and role-playing guidelines in a side panel, while the main panel provides a chat interface for interacting with the agent. 
Annotators send one message at a time while role-playing the specified user and terminate the interaction by issuing a \texttt{/stop} command once they believe the task has finished. 
After termination, the interface automatically presents a post-task survey in the chat interface to collect task-success judgments and interaction-quality ratings. 
The interface also exposes intermediate agent reasoning and tool-related traces (e.g., search or validation steps), which can occasionally make responses appear verbose when the agent performs multi-step operations, as reported by some annotators.
Our interface design aligns with popular AI agent interfaces, such as the ChatGPT interface, where users can interact with the agent in a chat interface and view extra information on the side.

\paragraph{Quality Control and LLM Judge Validation.}
We implemented an LLM judge to evaluate the quality of these interactions. 
We calibrated the LLM judge against a ground-truth set of 51 interactions that an author independently labeled ($N=51$). The confusion matrix is summarized in Table~\ref{tab:confusion_matrix}. 
\begin{table}[ht]
\centering
\caption{Confusion Matrix: LLM Judge vs.\ Human Ground Truth ($N=51$)}
\label{tab:confusion_matrix}
\renewcommand{\arraystretch}{1.2}
\setlength{\tabcolsep}{8pt}

\begin{tabular}{llccc}
\toprule
& & \multicolumn{2}{c}{\textbf{Human (Ground Truth)}} & \\
\cmidrule(lr){3-4}
& & \textbf{Fail} & \textbf{Pass} & \textbf{Total} \\
\midrule
\multirow{2}{*}{\textbf{LLM Judge}} 
& \textbf{Fail} & 9 (TN)  & 6 (FN)  & 15 \\
& \textbf{Pass} & 2 (FP)  & 34 (TP) & 36 \\
\midrule
\multicolumn{2}{l}{\textbf{Total}} 
& 11 & 40 & 51 \\
\bottomrule
\end{tabular}
\end{table}

While the Cohen's Kappa between human and LLM judge ($\kappa$=0.6) shows a moderate-to-substantial agreement, the LLM judge exhibits a conservative bias (low FP but high FN). 
Notably, the LLM judge achieved a high precision of $0.94$, indicating that 94\% of the interactions accepted by the LLM judge were confirmed as passing by human, and thus reasonably validate the integrity of the final dataset where we only keep the pass traces.

The LLM judge is powered by GPT-5 with the following prompt and all other parameters being default:
\begin{framed}

\small
You are a quality control system for annotation tasks. Your job is to assess whether the annotator:
\begin{enumerate}
    \item Role-played the user character well enough based on the instructions (The information in the \texttt{<|canvas|>} section is the instructions for the annotator to follow; the AI agent would not know these instructions. The interaction starts after the \texttt{<|canvas|>} section and ends after the USER posts the \texttt{/stop} command).
    \item Answered the survey responsibly and thoughtfully.
\end{enumerate}

\textbf{ANNOTATOR INSTRUCTIONS AND CONVERSATION:} \\
\texttt{\{interaction\_content\}}

Assess the quality on a scale of 0--100. Focus on:

\textbf{Role-Playing Quality:}
\begin{itemize}
    \item Did the annotator follow task instructions to complete the task with reasonable effort?
    \item Did the annotator follow character instructions (personality, goals, constraints) naturally?
    \item Did the annotator avoid hallucinating factual information not in the instructions (e.g., \texttt{user\_id})?
\end{itemize}

\textbf{Survey Response Quality:}
\begin{itemize}
    \item Do responses reflect the actual conversation experience?
    \item Do they show authentic human responses rather than AI-generated patterns?
\end{itemize}

Return JSON only:
\begin{verbatim}
{
  "score": <0-100>, // 80 is the passing score
  "reasoning": "<2-3 sentence explanation>",
  "flags": ["flag1", "flag2"],
  "is_spam": <true|false>
}
\end{verbatim}
\end{framed}

\subsection{Model details}
\label{app:model_details}

We evaluate 31 LLM user simulators grouped into three categories.
For each model, we evaluate it on 165 tasks in the $\tau$-bench dataset and ran 3 independent batches. We use the mean scores across batches as the final score for each model to compare with human annotators.

\paragraph{Proprietary (18 models).}
\textbf{GPT family:} GPT-3.5-turbo, GPT-4o, GPT-4o-mini, GPT-5-mini, GPT-5, GPT-5.1.
\textbf{Claude family:} Claude-3-Haiku, Claude-3.5-Sonnet, Claude-3.7-Sonnet, Claude-4.5-Haiku, Claude-Sonnet-4, Claude-Opus-4.
\textbf{Gemini family:} Gemini-2.0-Flash, Gemini-2.5-Flash, Gemini-2.5-Flash-Lite, Gemini-3-Flash, Gemini-3-Pro, Gemini-3.1-Pro.
All proprietary models are accessed via their respective APIs with default parameters.

\paragraph{Open-source (9 models).}
DeepSeek-V3.1, Llama-3.3-70B, Llama-4-Maverick, Qwen2.5-7B, Qwen3-235B, Qwen3-Next-80B, GPT-oss-120B, MiniMax-M2.5, and Kimi-K2.5.

\paragraph{Specialized (4 models).}
These models are specifically fine-tuned for user simulation or human-like behavior:
\textbf{CoSER-8B}~\citep{wang2025coser} (\texttt{Neph0s/CoSER-Llama-3.1-8B}),
\textbf{UserLM-8B}~\citep{dou2025simulatorarena},
\textbf{HumanLike-7B}~\citep{Wang2025HumanVA} (\texttt{wangxieric/Human-Like-Qwen2.5-7B-Instruct}), and
\textbf{HumanLM-opinion}~\citep{wu2026humanlm} (\texttt{snap-stanford/humanlm-opinion}).

\subsection{Survey ($\tau$-bench)}
\label{app:survey_taubench}

Please fill out the survey below based on your experience with the agent.

\paragraph{Feedback survey.}
\begin{enumerate}[nosep]
    \item \textbf{Did the agent successfully complete your task?} (Think about whether your goal was fully achieved by the end of the conversation.)
    \begin{itemize}[nosep]
        \item No --- due to a policy issue, which the agent clearly explained
        \item No --- task failed
        \item Partially --- some progress
        \item Yes --- task completed
        \item Fully --- exceeded expectations
    \end{itemize}

    \item \textbf{How efficient was the agent in completing the task?} (Did the agent complete the task in a reasonable number of steps?)
    \begin{itemize}[nosep]
        \item Very inefficient --- too many steps
        \item Somewhat inefficient
        \item About right
        \item Very efficient
    \end{itemize}

    \item \textbf{How did the number of clarifying questions feel to you?}
    \begin{itemize}[nosep]
        \item Too few
        \item About right
        \item Too many
    \end{itemize}

    \item \textbf{How much time/effort did it take to answer the agent's clarifying questions?} (Estimate based on the clarification phase only.)
    \begin{itemize}[nosep]
        \item Low
        \item Medium
        \item High
    \end{itemize}

    \item \textbf{Does the agent feel human-like?} (For example, does communicating with the agent feel like interacting with a human customer service representative?)
    \begin{itemize}[nosep]
        \item No
        \item Partially
        \item Yes
    \end{itemize}

    \item \textbf{How smooth was the overall interaction during clarification?} (Think about pacing, when it chose to ask vs.\ act, and whether it felt natural.)
    \begin{itemize}[nosep]
        \item Not smooth
        \item OK
        \item Smooth
        \item Excellent
    \end{itemize}

    \item \textbf{Overall agent performance score (1--5)} (Rate the agent's overall performance on this task.)
    \begin{itemize}[nosep]
        \item 1 (Very poor)
        \item 2 (Poor)
        \item 3 (Acceptable)
        \item 4 (Good)
        \item 5 (Excellent)
    \end{itemize}

    \item \textbf{If you encounter similar problems in life, would you like to reuse this agent?}
    \begin{itemize}[nosep]
        \item Absolutely no
        \item No
        \item Maybe
        \item Yes
        \item Absolutely yes
    \end{itemize}

    \item \textbf{Provide specific examples of agent behavior.} Help us understand the details:
    (i)~What did the agent do well?
    (ii)~What specific errors occurred?
    (iii)~Which rule violations happened?
    (iv)~What information did it forget?
    (v)~Copy-paste problematic agent messages if helpful.

    \item \textbf{How could the agent improve?} What changes would make the agent more effective?
\end{enumerate}

\subsection{Details of operationalizing  metrics}
\label{app:behavioral_metric_definitions}

This subsection defines the behavioral metrics used in Table~\ref{tab:behavioral_divergence}. Metrics are computed on the \emph{user} turns only, after filtering non-conversational meta messages (e.g., logging tokens, survey tags, and explicit stop markers). For word-based metrics, we strip agent-side markup (e.g., tool/function traces) before whitespace tokenization. Unless otherwise noted, ``X\%'' denotes the percentage of user turns in an interaction that satisfy a criterion, averaged across interactions.

\paragraph{D1: Communication styles.}
\begin{itemize}[nosep]
    \item \textbf{Wds/trn}: mean number of words per user turn.
    \item \textbf{Short\%}: fraction of user turns with $\leq 3$ words.
    \item \textbf{Polite\%}: fraction of user turns containing a politeness marker (e.g., ``please'', ``thanks'', ``sorry'').
    \item \textbf{Dash\%}: fraction of user turns containing an em-dash or en-dash. We use this only as an observable LLM-stylistic punctuation rate, not as a semantic measure of formality.
    \item \textbf{Ack\%}: fraction of user turns that are acknowledgment-only (e.g., ``ok'', ``sure'', ``got it'') with no additional content.
    \item \textbf{VerbCV}: coefficient of variation of user-turn word counts within an interaction (\(\mathrm{std}/\mathrm{mean}\)); higher implies more variation in verbosity across turns.
    \item \textbf{Repeat\%}: fraction of interactions that contain a highly repeated 3-gram in user turns (any trigram appearing $>$5 times), averaged as a percentage.
    \item \textbf{IDConf\%}: fraction of interactions in which the user exhibits ``identity confusion'' by using agent-like service language (e.g., ``how may I help'', ``let me check'', ``for verification purposes'').
\end{itemize}

\paragraph{D2: Information pattern.}
\begin{itemize}[nosep]
    \item \textbf{FrntId\%}: \emph{front-loading} ratio---percentage of all user words in an interaction that occur in the first two user turns.
    \item \textbf{IDs/trn}: mean number of identifier-like strings per user turn, using a regex that matches high-entropy tokens (e.g., long alphanumerics, long digit runs, emails, and structured IDs).
    \item \textbf{Wds/trn}: repeated here for convenience as a baseline information-volume measure.
    \item \textbf{Open.Wds}: mean number of words in the \emph{first} user turn of an interaction.
\end{itemize}

\paragraph{D3: Clarification.}
We classify question turns using mutually exclusive regex rules with priority \textbf{pushback} $\rightarrow$ \textbf{clarification} $\rightarrow$ \textbf{information-seeking}.
\begin{itemize}[nosep]
    \item \textbf{Uncert\%}: fraction of user turns containing uncertainty/hedging (e.g., ``maybe'', ``not sure'', ``I think'').
    \item \textbf{Certn\%}: fraction of user turns containing certainty markers (e.g., ``definitely'', ``for sure'', ``without a doubt'').
    \item \textbf{Pushbk\%}: fraction of user turns that match pushback patterns (e.g., ``are you sure?'', ``you already asked'').
    \item \textbf{ClarfyQ\%}: fraction of user turns that match clarification patterns (e.g., ``what do you mean?'', ``can you clarify?'') and are not counted as pushback.
    \item \textbf{InfoQ\%}: fraction of user turns that match information-seeking patterns (e.g., ``what is the status?'', ``can you check...'') and are not counted as pushback or clarification.
\end{itemize}

\paragraph{D4: Error reaction.}
\begin{itemize}[nosep]
    \item \textbf{Emot\%}: fraction of user turns containing emotion/frustration markers (e.g., ``frustrated'', ``annoyed'', ``ugh'').
    \item \textbf{Accuse\%}: fraction of user turns containing accusatory or strongly negative language (e.g., ``useless'', ``unacceptable'', ``scam'').
    \item \textbf{Pivot\%}: fraction of user turns indicating a strategy change or alternative request (e.g., ``instead'', ``on second thought'', ``let's try...'').
\end{itemize}

\subsection{Measurement validation and limits}
\label{app:measurement_limits}

Our lexical and regex metrics are intended as transparent distributional indicators rather than complete semantic models. For example, the dash rate captures a conspicuous LLM stylistic marker, but an em-dash does not by itself establish formality. As a complementary validation, we apply the RoBERTa formality ranker of \citet{dementieva2023formality}: it independently rates the default simulator as more formal than humans (mean $P(\mathrm{formal})$: 0.71 vs.\ 0.59; 70\% vs.\ 52\% of turns classified as formal). The D1 conclusion therefore does not depend on treating punctuation as formality.

D4 requires particular care. Emotion, accusation, and pivot rates are computed over all user turns; we do not first verify that each matched turn immediately follows an agent error. Consequently, D4 measures the prevalence of error-reaction-like behaviors at the interaction level, not a causal response to a specific failure. Our qualitative cases establish face-valid examples but are selected to illustrate patterns, not sampled to estimate precision or recall. Human annotation of error-linked turns is needed to validate this construct directly.

The study design imposes additional limits. Our annotators are US-based, 89\% report English as their primary language, and 66\% identify as White; cultural and linguistic variation may change communication style and error reactions. We fix the human-facing agent to GPT-5.2. The Gemini-agent analysis preserves simulator rankings, but does not show that humans would behave identically with a different agent. Role-playing a customer also differs from pursuing a genuine need. We chose three annotations per task to cover more tasks and models, favoring breadth over depth; this supports aggregate comparisons but gives limited per-task uncertainty estimates. Finally, known $\tau$-bench task and reward-quality issues~\citep{zhu2025rigorous} can affect absolute values, although our paired human--simulator comparisons hold tasks fixed.

\subsection{Persona-prompting ablation}
\label{app:persona_ablation}

To test whether default prompting alone creates the gap, we compare the standard $\tau$-bench prompt with a findings-informed persona prompt for GPT-5-mini on 50 identical tasks (25 airline, 25 retail) with GPT-5.2 as the agent. The persona prompt asks the simulator to reveal information gradually, vary tone and formality, express uncertainty, ask clarifying questions, push back, and show frustration after errors.

The prompt improves politeness and front-loading, but overshoots uncertainty and dash use; emotion does not increase, pivoting worsens, D4 collapses, and calibration degrades. These mixed shifts echo evidence that profile augmentation alone is unreliable~\citep{dou2025simulatorarena}.

\begin{table}[t]
\centering
\small
\setlength{\tabcolsep}{3.5pt}
\caption{Persona prompting moves some metrics toward the same-task human reference but overshoots others, lowering overall fidelity.}
\label{tab:persona_ablation}
\begin{tabular}{lrrr}
\toprule
Metric & Human & Default & Persona \\
\midrule
Polite\% & 11.0 & 44.0 & 28.0 \\
Front-load\% & 32.0 & 43.0 & 38.0 \\
Clarification\% & 0.8 & 0.0 & 1.0 \\
Uncertainty\% & 6.0 & 7.0 & 13.0 \\
Dash\% & 1.7 & 59.0 & 73.0 \\
Emotion\% & 1.3 & 1.3 & 0.7 \\
Pivot\% & 12.0 & 8.0 & 6.0 \\
\midrule
D4 & -- & 73.6 & 50.1 \\
ECE$\downarrow$ & -- & 0.18 & 0.29 \\
USI$\uparrow$ & -- & 70.9 & 64.6 \\
\bottomrule
\end{tabular}
\end{table}

\subsection{Task Success vs.\ Binary Reward}
\label{app:contingency}

\begin{figure*}[t]
\centering
\includegraphics[width=\textwidth]{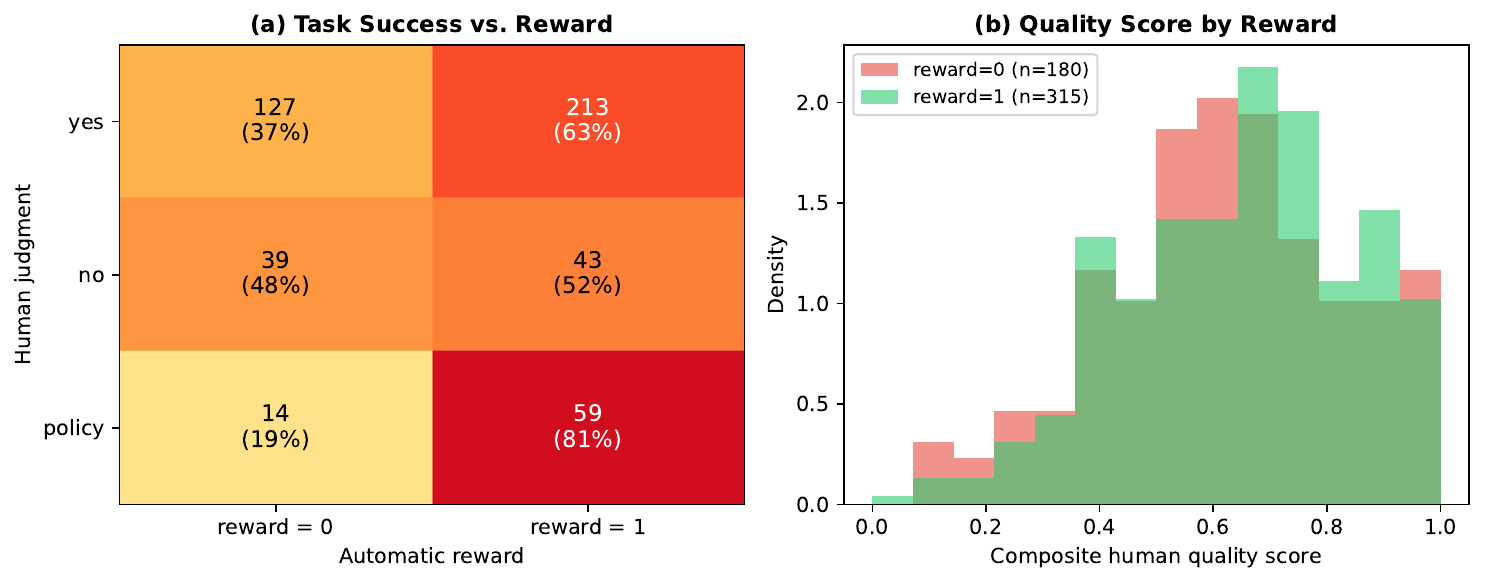}
\caption{(a)~Human task success judgment (Yes / No / Policy-constrained) vs.\ $\tau$-bench binary reward ($n{=}495$, Cram\'{e}r's $V = 0.168$). 70.6\% of reward{=}0 cases are judged successful by humans; 81\% of policy-constrained cases receive reward{=}1. (b)~Composite human quality scores by reward group. The binary reward predicts neither task success nor interaction quality as perceived by humans.}
\label{fig:contingency_composite}
\end{figure*}

Figure~\ref{fig:contingency_composite} shows the relationship between human task success judgments and $\tau$-bench binary rewards.

\begin{figure*}[t]
\centering
\includegraphics[width=\textwidth]{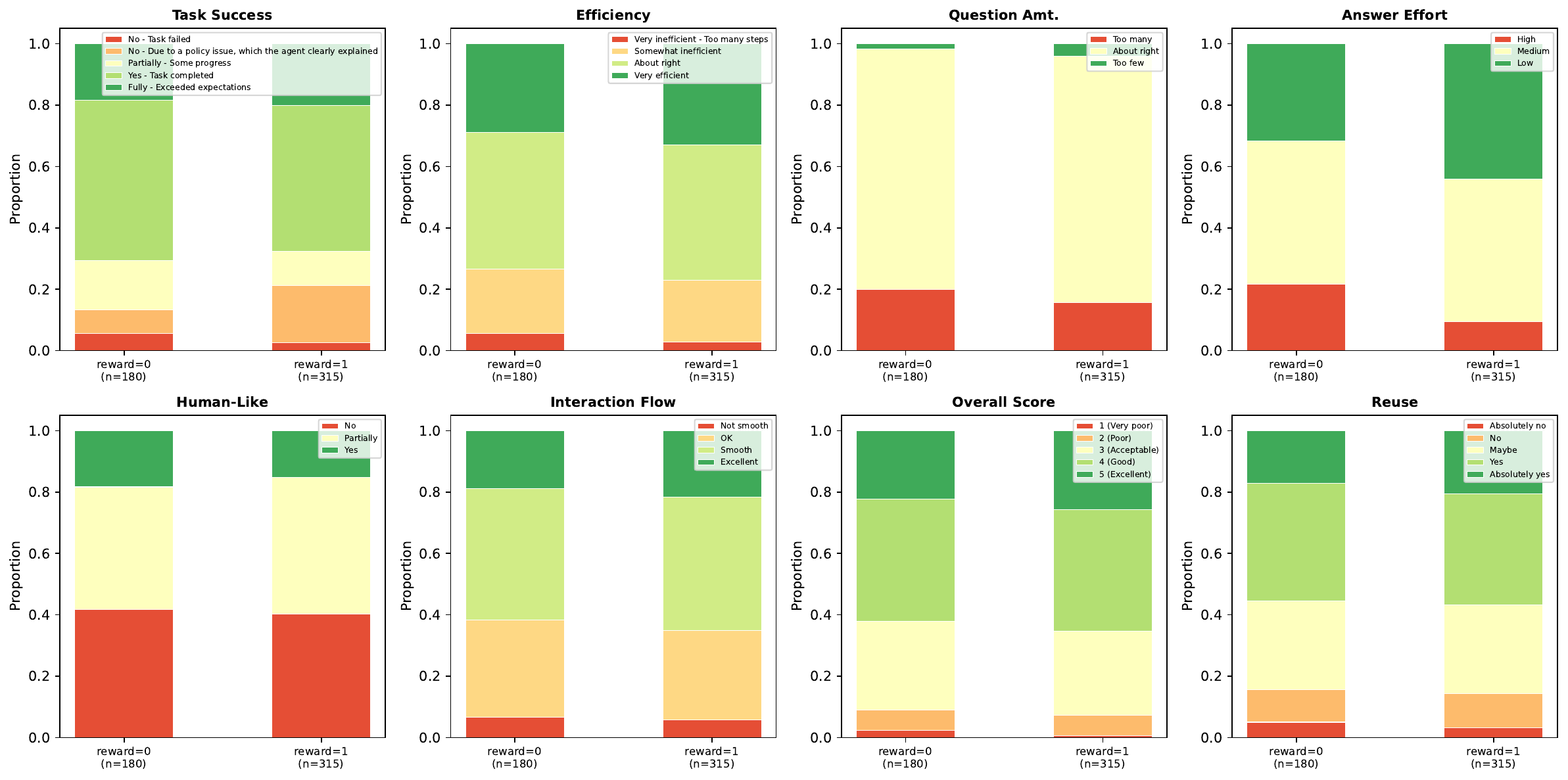}
\caption{Per-dimension human quality ratings by reward group ($n{=}495$). Stacked distributions for reward{=}0 and reward{=}1 are nearly indistinguishable across all eight dimensions, confirming that the binary reward captures none of these quality aspects.}
\label{fig:dimensions_by_reward}
\end{figure*}

\subsection{Agent Success Rates by User Simulator}
\label{app:success_rates}

\begin{figure*}[t]
\centering
\includegraphics[width=\textwidth]{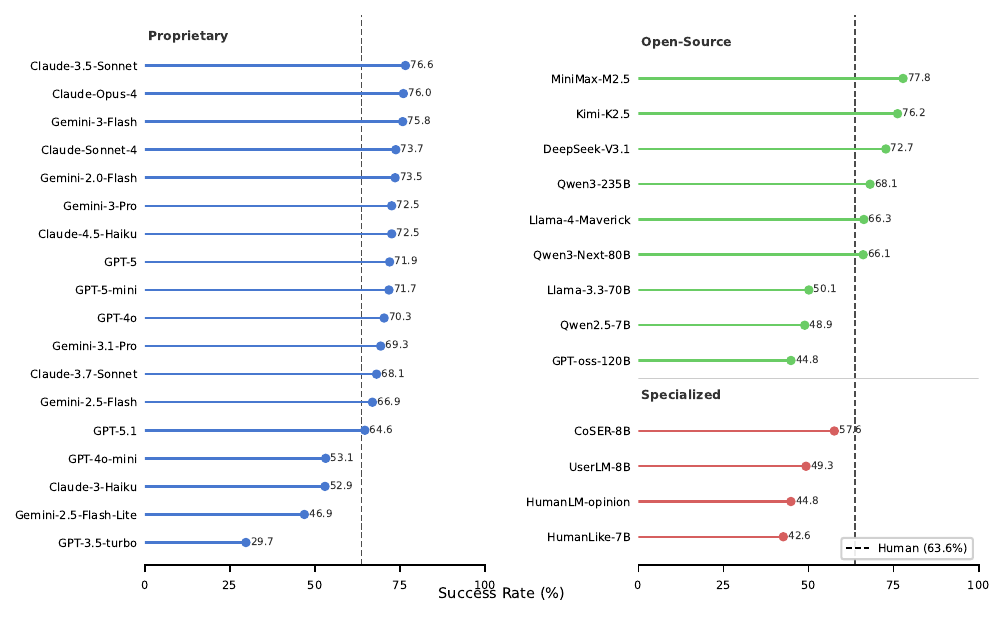}
\caption{Agent success rates (\%) on $\tau$-bench when paired with each user simulator. Models are grouped by category (proprietary, open-source, specialized). The dashed line indicates the human baseline. Most general-purpose LLM simulators inflate agent success rates above the human baseline, while specialized user-simulation models fall below it.}
\label{fig:success_rates}
\end{figure*}

Figure~\ref{fig:success_rates} shows agent success rates when paired with each user simulator.

\subsection{Example interactions}
\label{app:examples}

We present three case studies from $\tau$-bench comparing human users with GPT-4o as a representative LLM simulator on the same task instance. These examples illustrate the D1--D4 behavioral divergence patterns identified in \S\ref{sec:results_interactive}.

\paragraph{Case 1: Communication style and information pattern (D1, D2).}
In a retail return task (instance \texttt{retail\_48}), the human customer provides information tersely and incrementally, while the GPT-4o simulator front-loads details in polished, complete sentences.

\smallskip
\noindent\textit{Human user:}
\begin{quote}
\small
\textbf{Customer:} I need to return the air purifier I bought recently. It doesn't work as well as I'd like. \\
\textbf{Agent:} Please provide your name and ZIP code, plus the order ID if you have it. \\
\textbf{Customer:} Daiki Johnson 80273 \\
\textbf{Agent:} [authentication error---asks again] \\
\textbf{Customer:} daiki\_johnson\_9523 80273 \\
\ldots \\
\textbf{Customer:} I am not sure \hfill {\scriptsize (when asked to identify the order)}
\end{quote}

\noindent\textit{GPT-4o simulator (same task):}
\begin{quote}
\small
\textbf{Customer:} Hi there! I'd like to return an air purifier that I received. It's not working well. \\
\textbf{Agent:} Please provide your name and ZIP code, plus the order ID if you have it. \\
\textbf{Customer:} Sure! My name is Daiki Johnson and my ZIP code is 80273. I don't remember my order ID, unfortunately. \\
\textbf{Agent:} [finds the order immediately] \\
\textbf{Customer:} Yes, I confirm that I want to return only the air purifier from order \#W9502127, and I'd like the refund to be processed through PayPal. Thank you for handling this for me!
\end{quote}

\noindent The human gives minimal fragments (``Daiki Johnson 80273''), while the simulator produces full sentences with explicit context (``Sure! My name is\ldots and my ZIP code is\ldots''). The simulator also resolves the task in 3 turns; the human takes 8, revealing information only when prompted, a pattern consistent with the higher IDs-per-turn and lower short-turn rates we observe across models in D1 and D2.

\paragraph{Case 2: Error reaction (D4).}
In an airline modification task (instance \texttt{airline\_5}), the agent retrieves the wrong reservation. The human and simulator react very differently.

\smallskip
\noindent\textit{Human user:}
\begin{quote}
\small
\textbf{Customer:} \ldots the reservation should be 5RJ7UH. \\
\textbf{Agent:} [pulls up reservation 5RJ7UH: LAX$\to$SFO, not NY$\to$Chicago] \\
\textbf{Customer:} No, sorry, that's the wrong reservation. \\
\textbf{Agent:} Which reservation is your NY/IL trip? \\
\textbf{Customer:} UM3OG5 \\
\textbf{Agent:} [asks for confirmation of changes] \\
\textbf{Customer:} No \hfill {\scriptsize (blunt rejection)}\\
\ldots \\
\textbf{Customer:} Wrong reservation \hfill {\scriptsize (repeated twice)}
\end{quote}

\noindent\textit{GPT-4o simulator (same task):}
\begin{quote}
\small
\textbf{Agent:} UM3OG5 doesn't appear to be a New York to Chicago trip. It's a round trip SEA--DFW. \\
\textbf{Customer:} I'm sorry; I'm actually not sure which reservation ID is for the New York to Chicago trip. Could you help me identify it? As for UM3OG5, yes, please go ahead and add 3 bags to it. \\
\textbf{Agent:} [proceeds with bag addition] \\
\textbf{Customer:} Yes, please add the 3 checked bags to UM3OG5. \\
\textbf{Customer:} No preference, you can start with 5RJ7UH. \hfill {\scriptsize (cooperative pivot)}
\end{quote}

\noindent The human becomes terse and repetitive (``Wrong reservation''), signaling frustration without elaboration. The simulator apologizes, redirects politely, and even volunteers an alternative action (``please go ahead and add 3 bags'')---exhibiting the cooperative pivoting pattern captured by D4 metrics.

\paragraph{Case 3: Hedging and over-qualification (D3).}
In a retail exchange task (instance \texttt{retail\_101}), GPT-4o hedges extensively even when the task instructions are unambiguous.

\smallskip
\noindent\textit{Human user:}
\begin{quote}
\small
\textbf{Customer:} Um, hi. I actually need to return my luggage set, but I want to get the exact same one, just in red instead. Is that even possible? \\
\textbf{Customer:} Liam Thomas 85049 \\
\textbf{Customer:} I'm sorry I don't have the order ID, is that okay?
\end{quote}

\noindent\textit{GPT-4o simulator (same task):}
\begin{quote}
\small
\textbf{Customer:} I'm hoping to return a few items. \textbf{I'm not sure if this is even possible}, but I want to exchange my luggage set for the same model but in red. \\
\textbf{Customer:} Oh no, \textbf{maybe I'm cursed with bad luck today}! Alright, let's try with my email\ldots \textbf{I'm crossing my fingers this time}. \\
\textbf{Customer:} Ugh, can't anything go right? \textbf{Could it be my other email, maybe}? Try liam.t123@gmail.com. \textbf{Please work this time}!
\end{quote}

\noindent The human is direct and slightly tentative in one phrase (``Is that even possible?''). The simulator adds multiple hedges per turn (bolded), dramatizing uncertainty with phrases like ``I'm crossing my fingers'' and ``maybe I'm cursed.'' This over-hedging inflates the uncertainty rate captured in D3 and creates an artificially hesitant persona uncharacteristic of real customer interactions.

\subsection{Full Behavioral Divergence Results}
\label{app:full_results}

\begin{table*}[t]
\centering
\caption{Full behavioral divergence results across all 31 LLM user simulators on $\tau$-bench tasks (mean$\pm$std across three independent human annotation batches). \textbf{D1--D4} are per-dimension S{\o}rensen--Dice coefficients (higher$=$closer to human), see Appendix \ref{app:behavioral_metric_definitions} for definitions. \textbf{Eval} measures agreement between LLM and human evaluators on multi-dimensional surveys, computed as $(1{-}\mathrm{MAE})\times100$. The \textbf{User-Sim Index (USI)} aggregates D1--D4, Eval, and a calibration score $(1{-}\mathrm{ECE})\times100$ into a single 0--100 measure of human alignment.  Models are grouped into \colorbox{blue!5}{proprietary}, \colorbox{green!5}{open-source}, and \colorbox{orange!5}{specialized} categories, ranked by USI within each group. \textbf{Bold} indicates the best model per column.}
\label{tab:behavioral_divergence_full}
\small
\resizebox{\textwidth}{!}{
\begin{tabular}{@{}l cccc c cc c@{}}
\toprule
\textbf{Model} & \textbf{D1 Comm.} & \textbf{D2 Info.} & \textbf{D3 Clarif.} & \textbf{D4 React.} & \textbf{Eval} & \textbf{ECE}$\downarrow$ & \textbf{USI} \\
\midrule
Human (inter-ann.) & 87.4$_{\pm6.8}$ & 97.9$_{\pm0.9}$ & 88.0$_{\pm1.3}$ & 93.5$_{\pm2.5}$ & 97.4$_{\pm5.0}$ & 0.081$_{\pm0.010}$ & 92.7$_{\pm1.1}$ \\
\midrule
\rowcolor{blue!5} Gemini2.0-Flash & 51.6$_{\pm1.6}$ & \textbf{88.9$_{\pm1.1}$} & 68.2$_{\pm2.1}$ & 76.9$_{\pm3.7}$ & 73.7$_{\pm0.8}$ & 0.111$_{\pm0.022}$ & 74.7$_{\pm0.8}$ \\
\rowcolor{blue!5} GPT-5.1 & 47.3$_{\pm6.9}$ & 77.4$_{\pm0.6}$ & 73.3$_{\pm2.0}$ & \textbf{88.1$_{\pm2.6}$} & 72.1$_{\pm1.5}$ & 0.172$_{\pm0.011}$ & 73.5$_{\pm1.0}$ \\
\rowcolor{blue!5} GPT-5 & 49.7$_{\pm5.6}$ & 73.7$_{\pm0.7}$ & 73.2$_{\pm2.3}$ & 73.4$_{\pm3.3}$ & 74.5$_{\pm1.1}$ & 0.102$_{\pm0.022}$ & 72.4$_{\pm1.8}$ \\
\rowcolor{blue!5} GPT-4o-mini & 40.6$_{\pm1.8}$ & 84.7$_{\pm0.9}$ & 70.2$_{\pm3.7}$ & 73.7$_{\pm1.4}$ & 75.7$_{\pm0.3}$ & 0.123$_{\pm0.024}$ & 72.1$_{\pm0.8}$ \\
\rowcolor{blue!5} GPT-5-mini & 39.4$_{\pm5.9}$ & 74.4$_{\pm0.7}$ & \textbf{83.1$_{\pm2.3}$} & 68.7$_{\pm1.6}$ & 73.5$_{\pm0.5}$ & 0.102$_{\pm0.018}$ & 71.5$_{\pm1.4}$ \\
\rowcolor{blue!5} GPT-4o & 31.5$_{\pm2.5}$ & 84.4$_{\pm0.7}$ & 74.6$_{\pm2.9}$ & 72.4$_{\pm2.0}$ & 73.7$_{\pm1.1}$ & 0.096$_{\pm0.023}$ & 71.2$_{\pm1.3}$ \\
\rowcolor{blue!5} Gemini-2.5-Flash-Lite & 47.9$_{\pm1.8}$ & 86.0$_{\pm1.2}$ & 73.2$_{\pm2.0}$ & 59.5$_{\pm3.2}$ & 69.6$_{\pm1.0}$ & 0.184$_{\pm0.023}$ & 69.6$_{\pm0.3}$ \\
\rowcolor{blue!5} Gemini-3-Pro & 40.1$_{\pm2.7}$ & 81.4$_{\pm0.7}$ & 65.5$_{\pm5.3}$ & 57.6$_{\pm3.1}$ & 73.8$_{\pm0.2}$ & 0.125$_{\pm0.022}$ & 67.6$_{\pm0.9}$ \\
\rowcolor{blue!5} Claude-3.5-Sonnet & 39.2$_{\pm1.1}$ & 76.9$_{\pm0.7}$ & 59.6$_{\pm5.2}$ & 59.3$_{\pm3.5}$ & 74.1$_{\pm0.7}$ & 0.129$_{\pm0.035}$ & 66.0$_{\pm0.8}$ \\
\rowcolor{blue!5} Claude-4.5-Haiku & 25.9$_{\pm0.8}$ & 73.4$_{\pm0.8}$ & 55.6$_{\pm5.1}$ & 59.0$_{\pm3.5}$ & 75.4$_{\pm0.6}$ & 0.104$_{\pm0.020}$ & 63.2$_{\pm0.5}$ \\
\rowcolor{blue!5} Gemini-3-Flash & 37.7$_{\pm3.5}$ & 77.4$_{\pm0.7}$ & 56.5$_{\pm4.9}$ & 43.9$_{\pm2.5}$ & 71.7$_{\pm1.3}$ & 0.129$_{\pm0.029}$ & 62.4$_{\pm1.1}$ \\
\rowcolor{blue!5} GPT-3.5-turbo & 39.9$_{\pm1.1}$ & 74.1$_{\pm0.6}$ & 58.5$_{\pm1.5}$ & 59.9$_{\pm2.1}$ & 73.9$_{\pm0.7}$ & 0.339$_{\pm0.035}$ & 62.1$_{\pm0.2}$ \\
\rowcolor{blue!5} Claude-Sonnet-4 & 48.9$_{\pm1.5}$ & 68.0$_{\pm0.6}$ & 47.0$_{\pm4.7}$ & 43.3$_{\pm2.6}$ & 76.1$_{\pm0.8}$ & 0.114$_{\pm0.021}$ & 62.0$_{\pm0.7}$ \\
\rowcolor{blue!5} Gemini-2.5-Flash & 38.4$_{\pm1.3}$ & 73.5$_{\pm0.8}$ & 56.0$_{\pm4.8}$ & 43.6$_{\pm2.7}$ & 68.8$_{\pm0.7}$ & 0.086$_{\pm0.018}$ & 61.9$_{\pm0.7}$ \\
\rowcolor{blue!5} Claude-3-Haiku & 22.1$_{\pm1.0}$ & 55.7$_{\pm0.6}$ & 72.1$_{\pm4.0}$ & 56.9$_{\pm3.3}$ & \textbf{78.3$_{\pm0.5}$} & 0.143$_{\pm0.023}$ & 61.8$_{\pm0.4}$ \\
\rowcolor{blue!5} Gemini-3.1-Pro & 44.1$_{\pm1.9}$ & 67.1$_{\pm0.6}$ & 48.9$_{\pm4.6}$ & 45.3$_{\pm2.7}$ & 75.1$_{\pm0.9}$ & 0.101$_{\pm0.020}$ & 61.7$_{\pm0.6}$ \\
\rowcolor{blue!5} Claude-3.7-Sonnet & 26.4$_{\pm1.0}$ & 71.2$_{\pm0.7}$ & 50.8$_{\pm4.8}$ & 48.9$_{\pm3.1}$ & 72.7$_{\pm1.3}$ & \textbf{0.081$_{\pm0.010}$} & 60.3$_{\pm0.4}$ \\
\rowcolor{blue!5} Claude-Opus-4 & 32.6$_{\pm1.3}$ & 71.9$_{\pm0.7}$ & 46.6$_{\pm4.5}$ & 44.9$_{\pm2.8}$ & 73.4$_{\pm0.4}$ & 0.139$_{\pm0.029}$ & 59.2$_{\pm0.8}$ \\
\midrule
\rowcolor{green!5} DeepSeek-V3.1 & 45.1$_{\pm2.7}$ & 86.6$_{\pm1.0}$ & 74.5$_{\pm1.7}$ & 87.6$_{\pm2.0}$ & 74.3$_{\pm0.5}$ & 0.119$_{\pm0.030}$ & \textbf{76.0$_{\pm1.5}$} \\
\rowcolor{green!5} Llama-4-Maverick & 48.8$_{\pm0.7}$ & 82.6$_{\pm0.8}$ & 78.3$_{\pm3.1}$ & 66.6$_{\pm2.3}$ & 76.7$_{\pm1.0}$ & 0.107$_{\pm0.022}$ & 73.7$_{\pm1.0}$ \\
\rowcolor{green!5} Qwen3-235B & \textbf{60.8$_{\pm1.8}$} & 75.3$_{\pm0.6}$ & 71.5$_{\pm5.3}$ & 56.3$_{\pm2.4}$ & 74.6$_{\pm0.7}$ & 0.117$_{\pm0.014}$ & 71.1$_{\pm0.7}$ \\
\rowcolor{green!5} Qwen2.5-7B & 35.2$_{\pm1.3}$ & 70.9$_{\pm0.7}$ & 75.4$_{\pm6.1}$ & 74.8$_{\pm2.0}$ & 73.3$_{\pm1.2}$ & 0.125$_{\pm0.019}$ & 69.5$_{\pm1.5}$ \\
\rowcolor{green!5} Qwen3-Next-80B & 38.5$_{\pm1.5}$ & 73.2$_{\pm0.5}$ & 68.4$_{\pm2.7}$ & 67.9$_{\pm1.7}$ & 71.0$_{\pm2.2}$ & 0.087$_{\pm0.023}$ & 68.4$_{\pm0.8}$ \\
\rowcolor{green!5} GPT-oss-120B & 41.8$_{\pm1.1}$ & 63.9$_{\pm0.4}$ & 65.1$_{\pm0.9}$ & 77.0$_{\pm3.3}$ & 74.4$_{\pm1.3}$ & 0.155$_{\pm0.024}$ & 67.8$_{\pm0.6}$ \\
\rowcolor{green!5} MiniMax-M2.5 & 48.1$_{\pm0.4}$ & 72.8$_{\pm0.7}$ & 54.4$_{\pm5.3}$ & 62.1$_{\pm3.0}$ & 75.7$_{\pm0.1}$ & 0.188$_{\pm0.035}$ & 65.7$_{\pm0.7}$ \\
\rowcolor{green!5} Llama-3.3-70B & 43.9$_{\pm1.7}$ & 66.1$_{\pm0.6}$ & 54.1$_{\pm5.2}$ & 61.6$_{\pm2.5}$ & 74.6$_{\pm0.8}$ & 0.131$_{\pm0.035}$ & 64.6$_{\pm0.4}$ \\
\rowcolor{green!5} Kimi-K2.5 & 46.5$_{\pm1.9}$ & 66.0$_{\pm0.4}$ & 51.8$_{\pm5.3}$ & 47.2$_{\pm2.7}$ & 75.0$_{\pm1.1}$ & 0.123$_{\pm0.025}$ & 62.4$_{\pm0.7}$ \\
\midrule
\rowcolor{orange!5} CoSER-8B & 37.8$_{\pm1.1}$ & 71.5$_{\pm0.6}$ & 71.6$_{\pm2.0}$ & 69.9$_{\pm1.6}$ & 63.3$_{\pm0.6}$ & 0.109$_{\pm0.037}$ & 67.2$_{\pm0.5}$ \\
\rowcolor{orange!5} UserLM-8B & 30.8$_{\pm0.6}$ & 50.8$_{\pm0.4}$ & 56.8$_{\pm1.5}$ & 80.0$_{\pm3.5}$ & 67.4$_{\pm0.8}$ & 0.140$_{\pm0.030}$ & 62.0$_{\pm0.4}$ \\
\rowcolor{orange!5} HumanLike-7B & 35.7$_{\pm0.3}$ & 55.0$_{\pm0.5}$ & 51.6$_{\pm3.3}$ & 65.9$_{\pm3.4}$ & 72.8$_{\pm1.3}$ & 0.220$_{\pm0.029}$ & 59.8$_{\pm0.1}$ \\
\rowcolor{orange!5} HumanLM-opinion & 30.1$_{\pm0.5}$ & 19.5$_{\pm0.2}$ & 38.5$_{\pm5.6}$ & 50.7$_{\pm5.9}$ & 61.6$_{\pm0.3}$ & 0.192$_{\pm0.053}$ & 46.9$_{\pm0.7}$ \\
\bottomrule
\end{tabular}}
\end{table*}

\subsection{Robustness to Agent Choice}
\label{app:agent_robustness}

Throughout the main experiments we fix the agent to GPT-5.2 for controlled comparison. To test whether USI rankings generalize across agents, we repeat the evaluation with \textbf{Gemini-3.1-Pro} as the agent on five user simulators spanning proprietary models of varying capability: GPT-3.5-turbo, GPT-4o, GPT-5-mini, Gemini-2.0-Flash, and Gemini-3-Flash. Each simulator completes all 165 $\tau$-bench tasks (1 run) with the Gemini-3.1-Pro agent, and we compute the same D1--D4 behavioral metrics, ECE, and USI as in the main evaluation.

Table~\ref{tab:behavioral_divergence_gemini} reports the results. Because the human baseline was collected with the GPT-5.2 agent, the absolute USI values are not directly comparable to Table~\ref{tab:behavioral_divergence}---the agent influences conversation dynamics (e.g., how many clarification questions the agent asks, how it handles errors), which in turn affects user behavior. Nevertheless, the \emph{relative ordering} of user simulators is largely preserved: Gemini-2.0-Flash and GPT-4o remain the top-ranked simulators, while GPT-3.5-turbo remains near the bottom. This suggests that USI captures intrinsic properties of the user simulator rather than artifacts of a particular agent, supporting the generalizability of our rankings.

\begin{table}[t]
\centering
\caption{Behavioral divergence when using \textbf{Gemini-3.1-Pro} as the agent instead of GPT-5.2. Human baseline scores are from the GPT-5.2 agent setting and thus not directly comparable (see text). D1--D4 and USI definitions follow Table~\ref{tab:behavioral_divergence}.}
\label{tab:behavioral_divergence_gemini}
\small
\resizebox{\textwidth}{!}{
\begin{tabular}{@{}l cccc c cc c@{}}
\toprule
\textbf{User Simulator} & \textbf{D1 Conv.} & \textbf{D2 Info.} & \textbf{D3 Clarif.} & \textbf{D4 React.} & \textbf{Eval} & \textbf{ECE} & \textbf{USI} \\
\midrule
Human (GPT-5.2 agent) & 87.4$_{\pm6.8}$ & 97.9$_{\pm0.9}$ & 88.0$_{\pm1.3}$ & 93.5$_{\pm2.5}$ & 97.4$_{\pm5.0}$ & 0.069$_{\pm0.022}$ & 92.9$_{\pm0.9}$ \\
\midrule
Gemini-2.0-Flash & 71.6$_{\pm6.3}$ & 88.4$_{\pm1.1}$ & 76.9$_{\pm2.4}$ & 75.6$_{\pm3.8}$ & 73.7$_{\pm0.8}$ & 0.113$_{\pm0.025}$ & 79.1$_{\pm0.3}$ \\
GPT-4o & 42.6$_{\pm4.0}$ & 87.3$_{\pm0.8}$ & 78.4$_{\pm1.1}$ & 78.0$_{\pm3.0}$ & 73.7$_{\pm1.1}$ & 0.121$_{\pm0.025}$ & 74.6$_{\pm1.4}$ \\
GPT-5-mini & 54.5$_{\pm6.6}$ & 71.4$_{\pm0.7}$ & 69.1$_{\pm3.8}$ & 67.8$_{\pm2.0}$ & 73.5$_{\pm0.5}$ & 0.115$_{\pm0.035}$ & 70.8$_{\pm0.3}$ \\
Gemini-3-Flash & 55.8$_{\pm6.9}$ & 76.1$_{\pm0.7}$ & 67.4$_{\pm5.5}$ & 44.0$_{\pm2.6}$ & 71.7$_{\pm1.3}$ & 0.156$_{\pm0.025}$ & 66.6$_{\pm1.1}$ \\
GPT-3.5-turbo & 49.6$_{\pm1.2}$ & 76.6$_{\pm0.7}$ & 59.5$_{\pm3.3}$ & 43.9$_{\pm0.6}$ & 73.9$_{\pm0.7}$ & 0.105$_{\pm0.025}$ & 65.5$_{\pm0.5}$ \\
\bottomrule
\end{tabular}}
\end{table}

\subsection{Comparison to related work}

Table~\ref{tab:compare} compares our work to prior studies on measuring the gap between LLM-based user simulators and real users.

\begin{table*}[t]
\centering
\small
\setlength{\tabcolsep}{3pt}
\caption{Comparison to prior work on measuring the gap between LLM-based user simulators and real users across domains, interaction types, evaluation scale, alignment metrics, downstream evaluation, and whether user satisfaction is modeled.}\label{tab:compare}
\rotatebox{90}{
\begin{tabular}{@{} l l c l l l c @{} }
\toprule
& Tasks (dataset) & Task-type & Scale & Behavior Metric & Eval Metric & Sat.? \\
\midrule
\citet{seshadri2026lost} &
$\tau$-Bench (retail) &
H--A &
$\sim$360 p / 1440 sess &
ECE; miscalibration &
Task success-gap &
No \\

\citet{Bougie2025SimUSERSU} &
RS &
SimUser--RS &
1000 agents &
View ratio; click; rating acc. &
RS metric rank/acc. &
Yes \\

\citet{Ferraro2024AgentBasedMM} &
Twitter (2020 US election) &
Offline &
2020 election dataset &
Ling. patterns; homophily &
Macro dynamics &
No \\

\citet{Zhang2025ExploringHT} &
Web search &
Search &
31 p / 296 sess &
BLEU/BERTScore &
Click/stop F1 &
No \\

\citet{Zhang2024USimAgentLL} &
Web search &
Search &
Public behavior dataset &
Query gen; click/stop pred. &
IR sim quality &
No \\

\citet{Ren2024BASESLW} &
Web search (WARRIORS) &
Search &
WARRIORS dataset &
Query/click consis. ($\sim$90\%) &
MRR/NDCG &
No \\

\citet{Wang2025HumanVA} &
TOD &
H--A &
4 pers. scenarios &
10-dim taxonomy &
Diff. analysis &
Yes \\

\citet{davidson2023user} &
TOD (Human2Bot) &
H--A &
1165 dlg &
GSR match; div. &
TOD eval &
No \\

\citet{Ivey2024RealOR} &
Dialogue &
Offline &
100k dlg; 1.2k ann &
21 ling. features &
Align factors &
No \\

\citet{Wang2023UserBS} &
RS &
Mixed &
40 p &
Sem. align.; believability &
Ranking gap &
Yes \\

\citet{Yoon2024EvaluatingLL} &
CRS &
RS &
5 tasks / 4 src. &
Dist.; pref.; div.; coh. &
Gap analysis &
Yes \\

\citet{naous2025flipping} &
Coding / math &
H--A &
25.9k conv. &
PPL; div.; intent; nat. &
Asm. score &
No \\

\citet{dou2025simulatorarena} &
Math / writing &
H--A &
909 conv.; 18 asst. &
Match; profile fidelity &
Human corr. &
Yes \\

Ours &
TOD  &
H--A &
165 tasks / 495 sess. &
D1--D4; USI &
ECE; 6-dim quality &
Yes \\
\bottomrule
\end{tabular}
}
\end{table*}

\end{document}